\documentclass{article}

\usepackage{microtype}
\usepackage{graphicx}
\usepackage{subfigure}
\usepackage{booktabs} 

\usepackage{hyperref}

\usepackage[accepted]{icml2024}

\usepackage{amsmath}
\usepackage{amssymb}
\usepackage{mathtools}
\usepackage{amsthm}
\usepackage{amsfonts,bm}

\usepackage[capitalize,noabbrev]{cleveref}

\theoremstyle{plain}
\newtheorem{theorem}{Theorem}[section]

\newtheorem{lemma}[theorem]{Lemma}

\theoremstyle{definition}
\newtheorem{definition}[theorem]{Definition}

\theoremstyle{remark}

\usepackage{multirow}

\usepackage[textsize=tiny]{todonotes}

\icmltitlerunning{Layerwise Change of Knowledge in Neural Networks}

\begin{document}

\twocolumn[
\icmltitle{Layerwise Change of Knowledge in Neural Networks}



\icmlsetsymbol{equal}{*}

\begin{icmlauthorlist}
\icmlauthor{Xu Cheng}{equal,comp,yyy}
\icmlauthor{Lei Cheng}{equal,yyy}
\icmlauthor{Zhaoran Peng}{yyy}
\icmlauthor{Yang Xu}{xxx}
\icmlauthor{Tian Han}{zzz}
\icmlauthor{Quanshi Zhang\textsuperscript{§}}{yyy}
\end{icmlauthorlist}

\icmlaffiliation{yyy}{Shanghai Jiao Tong University.}
\icmlaffiliation{comp}{Nanjing University of Science and Technology.}
\icmlaffiliation{xxx}{Zhejiang University.}
\icmlaffiliation{zzz}{Stevens Institute of Technology}

\icmlcorrespondingauthor{Quanshi Zhang. Quanshi Zhang is the corresponding author. He is with the Department of Computer Science and Engineering, the John Hopcroft Center, at the Shanghai Jiao Tong University, China.}{zqs1022@sjtu.edu.cn}

\icmlkeywords{Machine Learning, ICML}

\vskip 0.3in
]



\printAffiliationsAndNotice{\icmlEqualContribution} 

\begin{abstract}
This paper aims to explain how a deep neural network (DNN) gradually extracts new knowledge and forgets noisy features
through layers in forward propagation. 
Up to now, although the definition of knowledge encoded by the DNN has not reached a consensus, \citet{pmlr-v202-li23at,Ren_2023_CVPR,ren2023we} have derived a series of mathematical evidence to take interactions as symbolic primitive inference patterns encoded by a DNN. 
We extend the definition of interactions and, for the first time, extract interactions encoded by intermediate layers. 
We quantify and track the newly emerged interactions and the forgotten interactions in each layer during the forward propagation, which shed new light on the learning behavior of DNNs. 
The layer-wise change of interactions also reveals the change of the generalization capacity and instability of feature representations of a DNN.
\end{abstract}

\section{Introduction}
\label{sec:intro}

Recently, understanding the black-box representation of deep neural networks (DNNs) has received increasing attention.
This paper investigates how a DNN gradually extracts knowledge from the input for inference during the layer-wise forward propagation, although the definition of~\textit{knowledge} encoded by an AI model is still an open problem.
To this end, the information bottleneck theory~\citep{shwartz2017opening,michael2018on} uses mutual information between the input and the intermediate-layer feature to measure knowledge encoded in each layer.
It finds that the DNN fits (learns) task-relevant information, and compresses task-irrelevant information.
\citet{Liang2020Knowledge} extract common feature components shared by different features as the shared knowledge.

\begin{figure*}[ht]
	\centering
	\includegraphics[width=\linewidth]{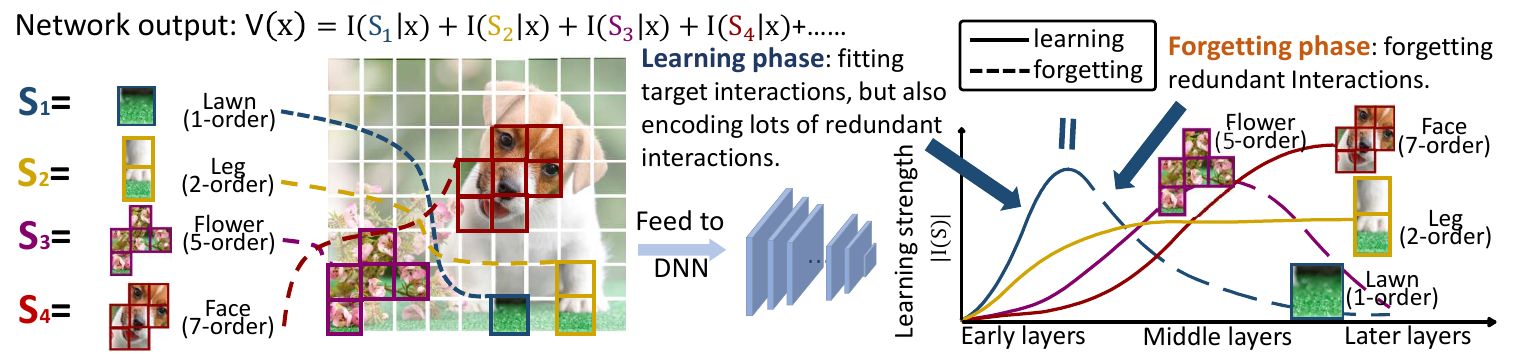}
	\vspace{-5pt}
	\caption{Tracking interactions through layers in the DNN. In most DNNs, early and middle layers usually fit target interactions modeled by the entire network at the cost of encoding lots of redundant interactions, and later layers remove such redundant interactions.}
	\label{fig:fig1}
\end{figure*}

In this paper, we aim to define and quantify the knowledge encoded in each layer. In this way, we can accurately decompose and track explicit changes of knowledge (i.e., the learning of new knowledge and the forgetting of old knowledge) through different layers.

However, there is no a widely-accepted definition of knowledge, because we cannot mathematically define/formulate knowledge in human cognition. 
Instead of focusing on cognitive issues, \citet{Ren_2023_CVPR,pmlr-v202-li23at} have discovered and \citet{ren2023we} have theoretically proven\footnote{\citet{ren2023we} have proven that the sparsity of interactions can be guaranteed by three common conditions for the DNN's smooth inferences on randomly masked samples.} \textbf{the sparsity property} and \textbf{universal-matching property} of interactions, \textit{i.e., given an input sample {\small $\boldsymbol{x}$}, a well-trained DNN usually only implicitly encodes a small number of interactions between the input variables, and the inference score can be explained as numerical effects of these interactions.}
\textbf{Thus, these two properties mathematically make such interactions (also called \textit{interaction primitives} or \textit{interaction concepts}) be considered as the knowledge encoded by a DNN.}
As Fig.~\ref{fig:fig1} shows, given a dog image {\small $\boldsymbol{x}$}, each interaction implicitly encoded by the DNN represents a co-appearance relationship between input variables (image patches) in {$S =\{\text{eye}, \text{nose}, \text{mouth}\}$}. This is actually an AND relationship between image patches in image {\small $\boldsymbol{x}$}.
Only when all patches in {\small$S$} are present in the image, the interaction {\small$S$} is activated and makes a numerical effect {\small$I(S|\boldsymbol{x})$} on the classification score.
Masking\footnotemark[4] any patch will deactivate the interaction {\small$S$} and remove the effect.

Although the above studies make it plausible to define and quantify interactions encoded by a DNN, our target of quantifying and tracking the interactions encoded by different layers presents the following three new challenges.
\\
(1)
\textbf{Alignment of interaction primitives.} The fair comparison between any arbitrary pair of layers requires interaction primitives extracted from different layers to be aligned, although the physical feature dimensions in different layers do not have a clear correspondence/alignment.
\\
(2)
\textbf{Decomposability and countability of knowledge.} Interactions help us overcome the challenge of representing uncountable knowledge as countable primitive patterns. In this way, we can exactly quantify how many interaction primitives are newly emerged and forgotten in each layer.
\\
(3) 
\textbf{Connection to the generalization capacity.} 
We hope to provide deep insights into how newly merged interaction primitives and forgotten old interaction primitives are related to the generalization capacity of a DNN.

Therefore, considering above challenges, we extend the definition of interactions to intermediate layers of a DNN. Specifically, given features of a certain layer, we train a linear classifier\footnote{
\citet{belinkov2022probing} discussed techniques and limitations of classifier probes. Please See Appendix~\ref{app_sec:discuss_probe} for the solutions to these problems.
} to use these features for classification, and extract a set of interactions from the classifier. We analyze the faithfulness of the newly proposed interaction towards the intermediate layers of a DNN, and we discover that the new interactions provide us with a more straightforward way to analyze how knowledge changes in the layerwise forward propagation. Instead of directly aligning features in different layers, we find that adjacent layers in a DNN usually encode similar sets of interactions. Thus, as illustrated in Fig.~\ref{fig:fig1}, we can clarify the emergence of new interactions and the forgetting of old interactions in each layer.

\textbf{Faithfulness of interactions.} More crucially, the newly defined interaction primitives still belong to the typical paradigm of interactions, so that there are a series of theorems~\citep{Ren_2023_CVPR,pmlr-v202-li23at,ren2023we} as convincing evidence to take countable/symbolic interactions as primitive inference patterns to represent uncountable knowledge in a DNN. 
Please See Section~\ref{sec:related} and Section~\ref{sec:concept} for details. 

In this way, we can use interactions to explain the change of the representation capacity of features in different layers from the following two perspectives, which can help both theoreticians and practitioners gain new insights into the learning behavior of a DNN.

$\bullet$\;
\textbf{The tracking of countable interactions in different layers reveals the change of representation complexity over different layers.}
The complexity of an interaction {\small$S$} is defined as the number of input variables in {\small$S$}, which is also termed the \textit{order} of this interaction,~\textit{i.e.,} {\small $\textit{order}(S)=\vert S \vert$}.
In experiments, 
we discover that in most DNNs, early and middle layers are usually trained to
fit target interactions encoded by the entire network at the cost of encoding lots of redundant interactions, and later layers remove such redundant interactions.

$\bullet$\;
\textbf{
Redefining the generalization capacity of DNNs and tracking generalizable interactions.}
The use of interaction primitives enables us to redefine the generalization power of a DNN from a new perspective.
That is, given multiple DNNs trained for the same task, if these DNNs encode similar interactions, then we consider interactions shared by different DNNs generalizable.
We discover that low-order interactions usually have stronger generalization capacity than high-order interactions.
Besides, we also discover that low-order interactions encoded by the DNN usually exhibit more consistent effects {\small$I(S|\boldsymbol{x}\prime=\boldsymbol{x}+\epsilon)$} when we add different small noises {\small$\epsilon$} to the input sample {\small$\boldsymbol{x}$}.
In comparison, high-order interactions often exhibit diverse effects {\small$I(S|\boldsymbol{x}\prime)$} on inference scores~\textit{w.r.t.} different noises {\small$\epsilon$}.
This indicates that low-order interactions often have higher stability.

Contributions of this study are summarized as follows. \\
(1) We redefine the interaction on intermediate layers, and find that the new definition ensures adjacent layers to encode similar interactions. \\
(2) Our study provides several theoretically verifiable metrics to quantify the newly emerged knowledge and forgotten knowledge in the forward propagation. \\
(3) The change of interactions is also found to be related to the generalization power of a DNN.

\section{Literature in Explaining Knowledge in DNNs}
\label{sec:related}

Explaining and quantifying the exact knowledge encoded by a DNN presents a significant challenge to explainable AI.
So far, there has not existed a widely accepted definition of knowledge that enables us to accurately disentangle and quantify knowledge encoded by intermediate layers of a DNN, because it covers multiple disciplinary issues, such as cognitive science, neuroscience,~\textit{etc}.
To explain and quantify the exact knowledge encoded by a  DNN,
 previous studies have either associated units of DNN feature maps with manually annotated semantics/concepts~\citep{bau2017network,kim2018interpretability} or automatically learned meaningful patterns from data~\citep{chen2019looks,  ijcai2021p409,zhang2020interpretable}, but they failed to provide a mathematically guaranteed boundary for the scope of each concept/knowledge. 
Thus, previous studies could not accurately quantify the exact amount of newly emerged/forgotten/unexplainable knowledge in each layer. Appendix~\ref{app_sec:related} provides further discussions of more methods~\citep{kolchinsky2018caveats, Liang2020Knowledge, michael2018on, shwartz2017opening, wang2022pacbayes}.

\textbf{Faithfulness of using interaction primitives to define knowledge in DNNs.}
Although there is no theory to guarantee that salient interactions can exactly fit the so-called \textit{knowledge} in human cognition, 
a series of studies have empirically verified and theoretically ensured the faithfulness of interaction primitives from the following perspectives.

(1)~\citet{pmlr-v202-li23at} have observed and \citet{ren2023we} have partially proven{\footnotemark[1]} that most DNNs encode a few interactions with salient effect $I(S|\boldsymbol{x})$ on the network output.

(2)~\citet{pmlr-v202-li23at} have observed that interactions exhibited considerable \textbf{generalization capacity} across samples and across models. Besides, they have also discovered that salient interactions exhibited remarkable \textbf{discrimination power} in classification tasks.

(3)~\citet{Ren_2023_CVPR} have proven seven desirable mathematical properties for interactions. 

(4) Interaction primitives can also be used to explain the representation capacity of DNNs.
\citet{deng2022discovering} have proven a counter-intuitive bottleneck of a DNN in encoding interaction primitives of the intermediate complexity.
\citet{liu2023towards} have proven the learning difficulty of interaction primitives.
\citet{zhou2023concept} have discovered that low-order interactions have higher generalization power than high-order interactions. 

Furthermore, we compare the interaction-based explanation  with attribution interpretability methods.
Please Appendix~\ref{app_sec:comp} for detailed discussions.


\section{Tracking Interactions through Layers}
\label{sec3}

\subsection{Preliminaries: using interactions to represent knowledge in DNNs}
\label{sec:concept}

So far, there is not a widely accepted way to define knowledge encoded by a DNN,
because the definition of knowledge is an interdisciplinary problem over cognitive science, neuroscience, and mathematics.
\citet{pmlr-v202-li23at}
has derived a series of properties as convincing evidence to define interactions as symbolic primitive inference patterns encoded by a DNN 
(please see Section~\ref{sec:related} for details).
Thus, in this paper, we extend the definition to quantify the change of interactions in the layer-wise forward propagation.
Specifically, there are two types of interactions, including AND interactions and OR interactions.

\begin{definition}[\textbf{AND interactions}]
\textit{Given an input sample {\small $\boldsymbol{x}=[x_1, x_2, \ldots, x_n]$} comprising {\small$n$} input variables, 
let {\small$N=\{ 1, 2, \ldots, n \}$} denote the indices of all {\small $n$} input variables, and let {\small$v(\boldsymbol{x})\in\mathbb{R}$} denote the scalar output of the DNN or a certain dimension of the DNN{\footnote{Note that people can apply different settings for {\small $v(\boldsymbol{x})$}. Here, we follow~\citep{deng2022discovering} to set 
{\small$v(\boldsymbol{x}) = \text{log} \frac{p(y=y^{\text{truth}} \vert \boldsymbol{x})}{1 - p(y=y^{\text{truth}} \vert \boldsymbol{x})} \in \mathbb{R}$} as the confidence of classifying the sample {\small$\boldsymbol{x}$} to the ground-truth category 
{\small$y^{\text{truth}}$}.}}.
Then, the AND interaction {\small$I_{\text{and}}(S \vert \boldsymbol{x})$} is used to quantify the effect of the AND (co-appearance) relationship among a subset {\small $S\subseteq N$} of input variables, which is encoded by the DNN {\small$v$} to compute the inference scores of the label $y^{\text{truth}}$.}
\begin{equation}\begin{small}\begin{aligned}
I_{\text{and}}(S \vert \boldsymbol{x}) = \sum\nolimits_{T \subseteq S} (-1)^{\vert S \vert - \vert T \vert} \cdot v(\boldsymbol{x}_T).
\label{eq:harsanyi_interaction_and}
\end{aligned}\end{small}\end{equation}
Here, {\small$\boldsymbol{x}_T$} denotes the masked\footnote{
We mask the input variable {\small$i \in N \setminus T$} to the baseline value {\small$b_i$} to represent its masked state. 
Here, we follow the widely-used setting of baseline values in~\citep{dabkowski2017real} to set {\small$b_i$} as the mean value of this variable across all samples in image classification, and follow \cite{shen2023inference} to set {\small$b_i$} as a special token (\textit{e.g.,} [MASK] token) in nature language processing.
Note that such settings of baseline values can bring in some biases~\cite{jain2022missingness}. To remove biases,~\citet{ren2023can} proposed a method to learn optimal
baseline values based on interactions.
Please see Appendix~\ref{app_sec:discuss_baseline_bias} for details.}
sample obtained by masking variables in {\small $N \setminus T$}, 
{\small$v(\boldsymbol{x}_T)$} represents the output score{\footnotemark[3]} for the target label $y^{\text{truth}}$ on the masked sample {\small$\boldsymbol{x}_T$}. 
\end{definition}

Each AND interaction with non-zero effect {\small$I_{\text{and}}(S \vert \boldsymbol{x}) \neq 0$} means that the DNN encodes the AND relationship between variables in {\small$S$}. The network output can be represented as the sum of interaction effects { \small$v(\boldsymbol{x})=\sum_{S\subseteq N}I_{\text{and}}(S \vert \boldsymbol{x})$}.

\textbf{OR interactions.}
\citet{Ren_2023_CVPR, zhou2023explaining} have further extended the AND interaction to the OR interaction. To this end, the overall network output is decomposed into the component for AND interactions {\small$v_\text{and}(\boldsymbol{x}_{T})$} and the component for OR interactions {\small$v_\text{or}(\boldsymbol{x}_{T})$}, subject to {\small $v_\text{and}(\boldsymbol{x}_{T})=0.5 \cdot v(\boldsymbol{x}_{T})+\gamma_T$} and {\small $v_\text{or}(\boldsymbol{x}_{T})=0.5\cdot v(\boldsymbol{x}_{T})-\gamma_T$}. {\small $\{\gamma_T\}$} is a set of learnable parameters to determine the decomposition. In this way, people can simultaneously explain AND interactions and OR interactions encoded by DNN. The AND interaction is extracted as {\small $I_{\text{and}}(S \vert \boldsymbol{x}) = \sum\nolimits_{T \subseteq S} (-1)^{\vert S \vert - \vert T \vert} \cdot v_{\text{and}}(\boldsymbol{x}_T$)}, just like in Eq.~(\ref{eq:harsanyi_interaction_and}). The OR interaction is defined as follows.

\begin{definition}[\textbf{OR interactions}]
\textit{The OR interaction is used to quantify the effect of the OR relationship between a set $S \subseteq N$ of input variables encoded by the DNN..}
\begin{equation}\begin{small}\begin{aligned}
I_{\text{or}}(S \vert \boldsymbol{x}) = -\sum\nolimits_{T \subseteq S} (-1)^{\vert S \vert - \vert T \vert} \cdot v_{\text{or}}(\boldsymbol{x}_{N\setminus T}).
\label{eq:harsanyi_interaction_or}
\end{aligned}\end{small}\end{equation}
\end{definition}

Eq.~\eqref{eq:harsanyi_interaction_or} indicates that
the presence of any input variable in {\small$S$} will activate the OR interaction and make an effect $I_{\text{or}}(S \vert \boldsymbol{x})$ to the output score {\small$v(\boldsymbol{x})$}.
\citet{Ren_2023_CVPR, zhou2023explaining} proposed to learn parameters {\small $\{\gamma_T\}$} to generate the sparsest AND-OR interactions. The AND-OR interactions is determined when {\small $\{\gamma_T\}$} are learned.
Please see \citet{zhou2023explaining} for detailed technique of learning {\small $\{\gamma_T\}$} for the optimal decomposition of AND-OR interactions.

\textbf{Faithfulness. The sparsity property and universal-matching property mathematically guarantee the faithfulness of interaction-based explanation.}
Let us randomly mask\footnotemark[4] an input sample {\small$x$} and generate a total of {\small $2^n$} masked samples {\small$\boldsymbol{x}_T$}. Then, Theorem \ref{them:matching} shows that output scores {\small$v(\boldsymbol{x}_T)$} on all {\small $2^n$} masked samples {\small$\boldsymbol{x}_T$} can always be well matched by AND-OR interactions.
\begin{theorem}\textbf{(Proven in Appendix~\ref{app_sec:proof_them1})}
\label{them:matching}
Given an input sample {\small $\boldsymbol{x}\in\mathbb{R}^{n}$}, the network output score {\small$v(\boldsymbol{x}_T)$} on each masked input samples {\small$\{\boldsymbol{x}_T\vert T\subseteq N\}$} can be decomposed into effects of AND interactions and OR interactions,
subject to {\small$I_{\text{and}}(\emptyset \vert \boldsymbol{x}) = v_\text{and}(\boldsymbol{x}_\emptyset)=v(\boldsymbol{x}_\emptyset)$} and {\small$I_{\text{or}}(\emptyset \vert \boldsymbol{x}) = v_\text{or}(\boldsymbol{x}_\emptyset)=0$}.
\begin{equation}\begin{small}\begin{aligned}
\label{eq:harsanyi_interaction_sum}
v(\boldsymbol{x}_T)
&= v_{\text{and}}(\boldsymbol{x}_T)+v_{\text{or}}(\boldsymbol{x}_T) \\
&= \sum\nolimits_{S\subseteq T} I_{\text{and}}(S|\boldsymbol{x}_T) + \sum\nolimits_{S\cap T \ne \emptyset} I_{\text{or}}(S|\boldsymbol{x}_T)
\end{aligned}\end{small}\end{equation}
\end{theorem}
\citet{ren2023we} have proven\footnotemark[1] that most AND-OR interactions have negligible effects {\small$I(S|\boldsymbol{x})\approx 0$} on inference, which can be regarded as noisy patterns. Only a small number of interactions have considerable effects. Given an input sample {\small $\boldsymbol{x}\in\mathbb{R}^{n}$}, \textbf{we can use a small set of salient AND interactions {\small$\Omega^{\text{and}}_{\text{salient}}$} and OR interactions {\small$\Omega^{\text{or}}_{\text{salient}}$} to universally match network outputs {\small$v(\boldsymbol{x}_T$)} on all {\small$2^n$} masked samples}. This indicates that salient interactions can serve as \textbf{primitive inference patterns} encoded by the DNN.
{\begin{lemma}\textbf{(Proving interactions as primitive inference patterns}, c.f. Appendix~\ref{app_sec:proof_lemma1}) 
\label{le:faithfulness}
Given an input sample {\small $\boldsymbol{x}\in\mathbb{R}^{n}$}, the network output on all {\small $2^n$} masked input samples {\small$\{\boldsymbol{x}_T\vert T\subseteq N\}$} can be universally matched by a small set of salient interactions.
\begin{equation}\begin{small}\begin{aligned}
v(\boldsymbol{x}_T) 
\approx v(\boldsymbol{x}_\emptyset) 
+ {\sum_{\substack{S \in \Omega^{\text{and}}_{\text{salient}}\\\emptyset \neq S\subseteq T}} {\hspace{-2mm} I_{\text{and}}(S\vert \boldsymbol{x_T})}}
+ {\sum_{\substack{S \in \Omega^{\text{or}}_{\text{salient}}\\S\cap T \ne \emptyset}}{\hspace{-2mm} I_{\text{or}}(S\vert \boldsymbol{x_T})}}
\end{aligned}\end{small}\end{equation}
\end{lemma}}

\begin{figure*}[t]
	\centering
	\includegraphics[width=\linewidth]{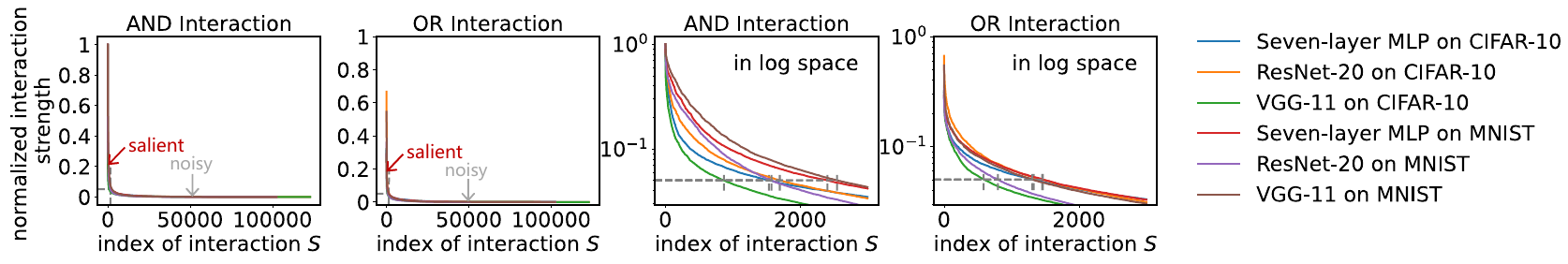}
	\vspace{-8pt}
	\caption{Sparsity of interactions.
	We visualized strength of all AND-OR interactions extracted from different samples {\small$\boldsymbol{x}$}, {\small$\vert I(S \vert \boldsymbol{x})\vert$} \textit{w.r.t.} different {\small$S$} and {\small$\boldsymbol{x}$}, in a descending order.
Only about $21.8$ AND/OR interactions in each sample of the MNIST dataset and about $45.6$ AND/OR interactions in each sample of the CIFAR-10 dataset made salient effects on the network output.}
	\label{fig:spa}
	\vspace{-3pt}
\end{figure*}

\subsection{Tracking interactions through layers}
\label{sec:3.2}

Despite the universal-matching property, the transferability, and the discrimination power of interactions in Section~\ref{sec:related}, the definition, quantification, and tracking of interactions through layers present distinctive challenges in real applications. Specifically, we aim to define the interaction for the high-dimensional features, while previous interactions are all defined on a scalar output score. Besides, the newly defined interaction successfully ensures that neighboring layers encode similar interactions. This enables us to propose various metrics to track the newly emerged interaction primitives and the forgotten interaction primitives in each layer, which provide new insights into the learning of DNNs.

\subsubsection{Verifying the sparsity of interactions}
\label{sec:vis}

Before we define interactions encoded by intermediate-layer features, we need to first examine whether the final layer of the DNN encodes a small number of interactions.
Although the sparsity of interactions has been partially proven under three common conditions{\footnotemark[2]}, it is still a challenge to strictly examine whether the DNN fully satisfies these conditions in real applications.
Besides, the sparsity of interactions has not been proven when we simultaneously use AND interactions and OR interactions to explain a DNN.

The interactions used by the final layer are directly extracted based on the network output score {\small$v(\boldsymbol{x})$}{\footnotemark[3]}, according to Eq.~(\ref{eq:harsanyi_interaction_and}) and Eq.~(\ref{eq:harsanyi_interaction_or}).
Thus, we can consider interactions extracted from the final layer as the target interactions used for the inference.
If these interactions are sparse, then the utility of all layers can be simplified as pushing features towards a specific small set of sparse interactions. This will significantly simplify feature analysis.

\textbf{\textit{Experiments.}}
We conducted experiments to illustrate the sparsity of interactions.
Given a well-trained DNN and an input sample {\small $\boldsymbol{x}\in\mathbb{R}^{n}$}, we calculated AND interactions {\small$I_{\text{and}}(S \vert \boldsymbol{x})$} and OR interactions {\small$I_{\text{or}}(S \vert \boldsymbol{x})$} of all {\small$2^{n}$} possible subsets\footnote{Appendix~\ref{app_sec:part} introduces the details of selecting a relatively small number of input variables (image patches or words) to compute interactions in order to reduce computational cost.} {\small$S \subseteq N$}.
To this end, we trained VGG-11~\citep{simonyan2014very}, ResNet-20~\citep{he2016deep} on the MNIST dataset~\citep{lecun1998gradient} and CIFAR-10 datasets~\citep{krizhevsky2009learning}, respectively.
We also learned a seven-layer MLP (namely \textit{MLP-7}) on the MNIST dataset and CIFAR-10 dataset, respectively, where each layer contained {$1024$} neurons.
Please see Appendix~\ref{app_sec:exp_sparsity} for experimental details.

Fig.~\ref{fig:spa} shows the strength of all AND-OR interactions extracted from different samples {\small$\boldsymbol{x}$}, {\small$\vert I(S \vert \boldsymbol{x})\vert$} \textit{w.r.t.} different {\small$S$} and {\small$\boldsymbol{x}$}, in descending order.
We discovered only about $21.8$ AND/OR salient interactions in each MNIST image and about $45.6$ AND/OR salient interactions in each CIFAR-10 image. All other interactions exhibited very small effects. Such a phenomenon verified the sparsity of interactions.

\subsubsection{Extracting interactions from intermediate layers}
\label{sec:3.2.2}

In comparison with extracting interactions from the network output score {\small$v(\boldsymbol{x})$}{\footnotemark[3]}, defining and extracting interactions from intermediate layers present a new challenge. 
It is because the intermediate-layer features are usually high-dimensional vectors/tensors/matrices, rather than a scalar output.
Thus, we need to define a new scalar metric {\small$v^{(l)}(\boldsymbol{x})$}, which faithfully identify signals in the high-dimensional feature directly related to the classification task, to compute interactions encoded by the $l$-th layer of the DNN.

To this end, given an input sample {\small $\boldsymbol{x}$}, we propose to train a linear classifier {\small $p^{(l)}(y|\boldsymbol{x})=\textit{\text{softmax}}/\textit{\text{sigmoid}}((w^{(l)})^T f^{(l)}(\boldsymbol{x}) +b^{(l)})$} based on the cross-entropy loss,
which uses the feature {\small$f^{(l)}(\boldsymbol{x})$} of the $l$-th layer to conduct the same classification task as the DNN\footnote{Appendix~\ref{app_sec:train} introduces the details of training the classifier. Note that the network parameters in the DNN are all fixed without being tuned, when we learn classifiers.}. We can define the following {\small $v^{(l)}(\boldsymbol{x})$} to represent signals encoded by the $l$-th layer of the DNN.
\begin{equation}\begin{small}\begin{aligned}
v^{(l)}(\boldsymbol{x}) &= \text{log} \frac{p^{(l)}(y=y^{\text{truth}} \vert \boldsymbol{x})}{1 - p^{(l)}(y=y^{\text{truth}} \vert \boldsymbol{x})}-\delta_{N}, \\
v^{(l)}(\boldsymbol{x}_{T}) &= \text{log} \frac{p^{(l)}(y=y^{\text{truth}} \vert \boldsymbol{x}_{T})}{1 - p^{(l)}(y=y^{\text{truth}} \vert \boldsymbol{x}_{T})}-{\delta}_{T},
\end{aligned}\end{small}\end{equation}
where {\small$\delta_T$} is a learnable residual proposed to model and remove the tiny noise from the output {\small$v^{(l)}(\boldsymbol{x}_{T})$}, so as to extract relatively clean interactions.
{\small$\delta_T$} is constrained to a small range {\small$\kappa=0.04\cdot\vert v^{(l)}(\boldsymbol{x}_{N})-v^{(l)}(\boldsymbol{x}_{\emptyset})\vert$}.
We discover that small noise in output function {\small$v^{(l)}(\boldsymbol{x}_T)$} may significantly change the interaction effect.
In this way, parameters {\small$\{\gamma_T, \delta_T\}$} are learned by minimizing {\small$\sum\nolimits_{T\subseteq N} \vert I_\text{and}(T\vert\boldsymbol{x},v^{(l)})\vert+\vert I_\text{or}(T\vert\boldsymbol{x},v^{(l)})\vert, s.t.\;\forall T\subseteq N, \vert \delta_T \vert < \kappa$}.
An ablation study in Appendix~\ref{app_sec:ablation} shows that the extraction of interactions is relatively robust to the {\small $\kappa$} value.

\textbf{Comparing interaction complexity over different layers.}
The new function {\small$v^{(l)}(\boldsymbol{x})$} enables a fair comparison between interactions extracted from different layers.
The classification score {\small$v^{(l)}(\boldsymbol{x})$} potentially reflects a set of interactions, which are encoded by {\small$f^{(l)}(\boldsymbol{x})$} and can be  directly used for classification. 
Specifically, we conducted experiments to extract interactions from different layers of different DNNs\footnotemark[5]. We used the MLP-7, VGG-11, and ResNet-20 trained on the MNIST dataset and CIFAR-10 dataset, which were introduced in Section~\ref{sec:vis}.
We also fine-tuned pre-trained DistilBERT~\citep{sanh2019distilbert} and $\text{BERT}_\text{BASE}$~\citep{devlin-etal-2019-bert} models on the SST-2 dataset~\citep{socher-etal-2013-recursive} for binary sentiment classification.

\begin{figure*}[t]
	\centering
	\includegraphics[width=\linewidth]{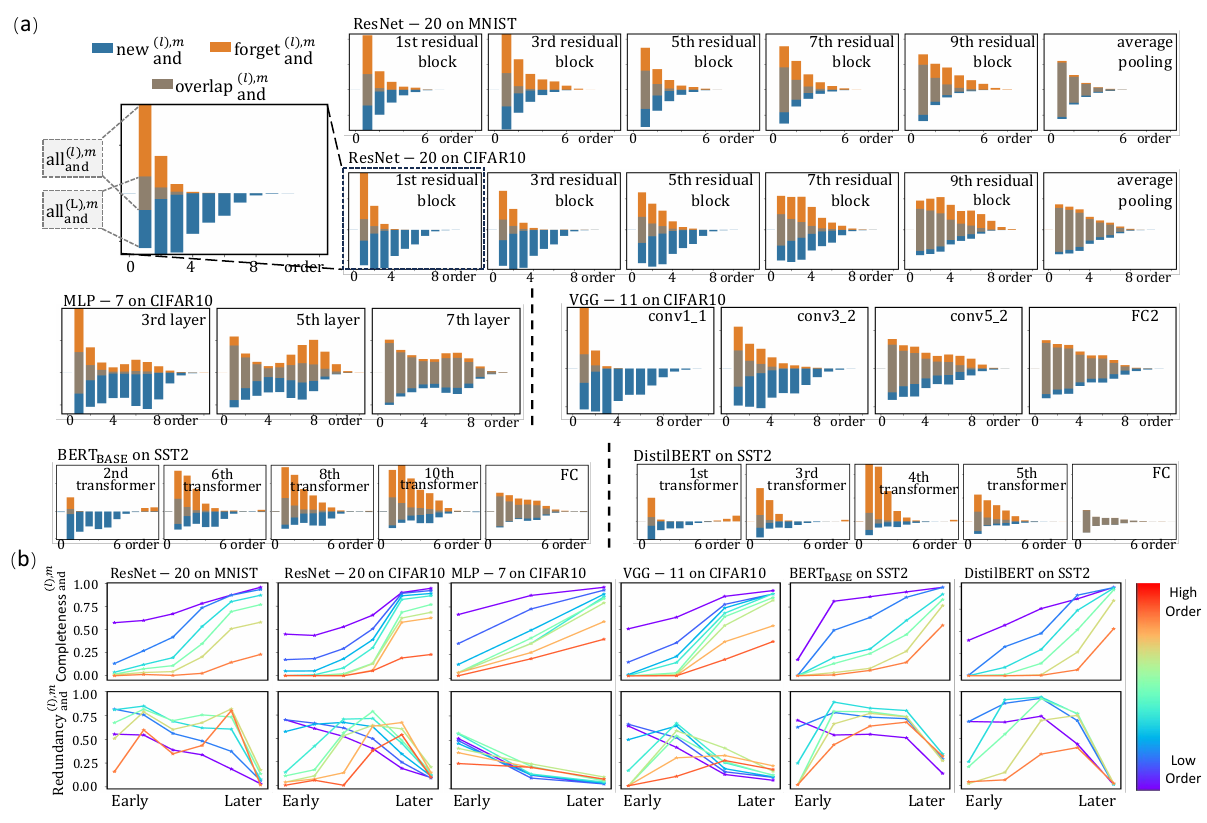}
	\vspace{-5pt}
	\caption{(a)~Tracking the change of the average strength of the overlapped ({\small$\text{\textit{overlap}}_{\text{and}}^{(l), m}$}), forgotten ({\small$\text{\textit{forget}}_{\text{and}}^{(l), m}$}), and newly emerged interactions ({\small$\text{\textit{new}}_{\text{and}}^{(l), m}$}) through different layers. For each subfigure, the total length of the orange bar and the grey bar equals to {\small$\textit{\text{all}}_\text{and}^{(l), m}$}, and the total length of the blue bar and the grey bar equals to {\small$\textit{\text{all}}_\text{and}^{(L), m}$}
	(b)~Tracking the change of {\small $\textit{completeness}^{(l), m}_{\text{and}}$} and {\small $\textit{redundancy}^{(l), m}_{\text{and}}$} through different layers. We do not show interactions of the highest four orders, because almost no interactions of extremely high orders were learned.
	Please see Appendix~\ref{app_sec:more_result} for results of OR interactions and results on tabular datasets.}
	\label{fig:overlap}
\end{figure*}

We used the order of an interaction to measure the complexity of the interaction. The order was defined as the number of input variables involved in this interaction,~\textit{i.e.,}  {\small$\textit{order}(S)=\vert S \vert$}. As illustrated in Fig.~\ref{fig:overlap}, 
linear classifiers trained on features of early layers encode less high-order interactions 
than later layers. We can consider that features in early layers usually represent lots of local and simple non-linear patterns between a few input variables, but most of such patterns cannot be directly used by the classifier for the classification task. 
Besides, compared to linear classifiers trained on early-layer features,
classifiers trained on features of later layers usually  share more similar interactions with the final layer of the DNN.


\textbf{Emergence of new interactions and discarding of old interactions.}
In this experiment, we quantified how the DNN gradually learned new interactions and discarded useless interactions in the forward propagation and obtained the target interactions in the last layer.
To this end, given all AND-OR interactions encoded by the $l$-th layer, let {\small$\Omega_{\text{and}}^{(l),m}= \{S \subseteq N: \vert S\vert=m, \vert I_\text{and}(S \vert \boldsymbol{x}, v^{(l)} ) \vert > \tau\}$}\footnote{We set {\small $\tau=0.05 \cdot \max_{S}(\vert I_\text{and}(S\vert \boldsymbol{x},v^{(l)})\vert, \vert I_\text{or}(S\vert \boldsymbol{x},v^{(l)})\vert)$}
to select a set of salient interactions from all interactions extracted from the $l$-th layer of the target DNN.} denote the set of salient AND interactions of the $m$-th order extracted from the $l$-th layer.
Accordingly, {\small$\Omega_{\text{or}}^{(l),m}= \{S \subseteq N: \vert S\vert=m, \vert I_\text{or}(S \vert \boldsymbol{x}, v^{(l)} ) \vert > \tau{\footnotemark[7]} \}$} represented the set of salient OR interactions of the $m$-th order extracted from the $l$-th layer.
To this end, we used {\small$\textit{\text{all}}_\text{and}^{(l), m}$} and {\small$\textit{\text{all}}_\text{and}^{(L), m}$} to quantify the overall strength of all $m$-order salient AND interactions encoded by the $l$-th layer and those encoded by the final layer (the $L$-th layer), respectively.
\begin{equation}\begin{small}\begin{aligned}
\textit{\text{all}}_\text{and}^{(l), m} =  \sum\nolimits_{S\in \Omega_{\text{and}}^{(l),m}} \vert I_\text{and}(S \vert \boldsymbol{x}, v^{(l)})\vert, \\
\textit{\text{all}}_\text{and}^{(L), m} =  \sum\nolimits_{S\in \Omega_{\text{and}}^{(L),m}} \vert I_\text{and}(S \vert \boldsymbol{x}, v^{(L)})\vert.
\end{aligned}\end{small}\end{equation}
As Fig.~\ref{fig:overlap} shows, we designed the following three metrics to further disentangle the overall strength {\small$\textit{\text{all}}_\text{and}^{(l), m}$} and {\small$\textit{\text{all}}_\text{and}^{(L), m}$} into three terms:
(1) the overall strength of interactions shared by both the $l$-th layer and the final layer, {\small$\text{\textit{overlap}}_{\text{and}}^{(l), m}$},
(2) the overall strength of interactions encoded by the $l$-th layer but later forgotten in the final layer, {\small$\text{\textit{forget}}_{\text{and}}^{(l), m}$},
(3) the overall strength of interactions that were encoded in the final layer, but were not encoded by the $l$-th layer, {\small$\text{\textit{new}}_{\text{and}}^{(l), m}$}.
\begin{equation}\begin{small}\begin{aligned}
\text{\textit{overlap}}_{\text{and}}^{(l), m} &= \sum\nolimits_{S\in \Omega_{\text{and}}^{(l),m} \bigcap  \Omega_{\text{and}}^{(L),m}}\;
\left\vert I_{\text{and, shared}}^{(l, L)}(S\vert \boldsymbol{x})\right\vert,
\\
\text{\textit{forget}}_{\text{and}}^{(l), m} &= \sum\nolimits_{S\in \Omega_{\text{and}}^{(l),m}}
\left\vert I_{\text{and}}(S\vert \boldsymbol{x},v^{(l)}) - I_{\text{and, shared}}^{(l, L)}(S\vert \boldsymbol{x})\right\vert,
\\
\text{\textit{new}}_{\text{and}}^{(l), m} &= \sum\nolimits_{S\in \Omega_{\text{and}}^{(L),m}}
\left\vert  I_{\text{and}}(S\vert \boldsymbol{x},v^{(L)})  - I_{\text{and, shared}}^{(l, L)}(S\vert \boldsymbol{x})\right\vert,
\end{aligned}\end{small}\end{equation}
where {\small$I_{\text{and, shared}}^{(l, L)}(S\vert \boldsymbol{x})$} measured the shared AND interactions between {\small$I_{\text{and}}(S\vert \boldsymbol{x},v^{(l)})$} extracted from the $l$-th layer and {\small$I_{\text{and}}(S\vert \boldsymbol{x},v^{(L)})$} encoded by the final $L$-th layer.
If {\small$I_{\text{and}}(S\vert \boldsymbol{x},v^{(l)})$} and {\small$I_{\text{and}}(S\vert \boldsymbol{x},v^{(L)})$} had opposite interaction effects, then {\small$I_{\text{and, shared}}^{(l, L)}(S\vert \boldsymbol{x})=0$};
Otherwise, the shared AND interaction was defined as {\small$I_{\text{and, shared}}^{(l, L)}(S\vert \boldsymbol{x})=
\text{sign}(I_{\text{and}}(S\vert \boldsymbol{x},v^{(l)}))\cdot
\min(\vert I_{\text{and}}(S\vert \boldsymbol{x},v^{(l)})\vert,
\vert I_{\text{and}}(S\vert \boldsymbol{x},v^{(L)})\vert )$}.

\begin{figure*}[ht]
	\centering
	\includegraphics[width=\linewidth]{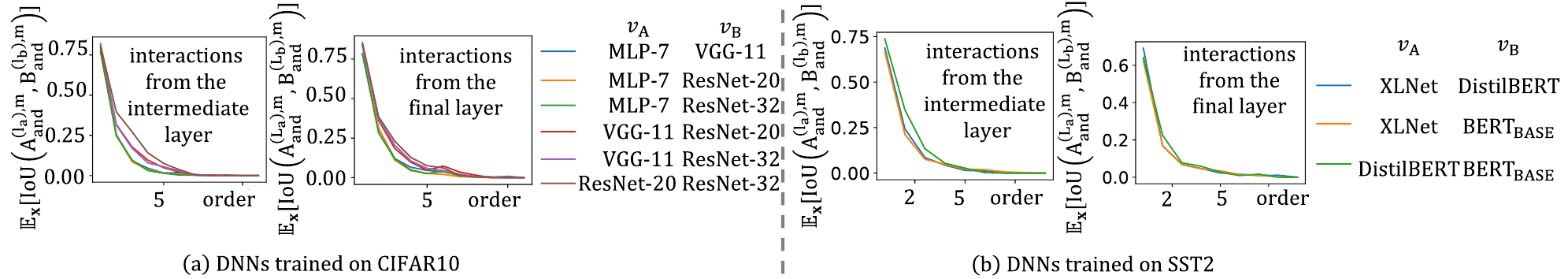}
	\caption{Average IoU values of AND interactions extracted from two DNNs trained for the same task over different input samples. Low-order interactions usually exhibited higher IoU values, thereby being better generalized across DNNs. Please see Appendix~\ref{app_sec:more_result} for results of OR interactions and Appendix~\ref{app_sec:exp_detail_layer} for the selected intermediate layer.}
	\label{fig:generalization}
 \end{figure*}
Thus, {\small$\text{\textit{overlap}}_{\text{and}}^{(l), m}$}, {\small$\text{\textit{forget}}_{\text{and}}^{(l), m}$}, and {\small$\text{\textit{new}}_{\text{and}}^{(l), m}$} formed a decomposition of overall interaction strength, as follows.

\begin{equation}\begin{small}\begin{aligned}
\textit{\text{all}}_\text{and}^{(l), m} =\text{\textit{overlap}}_{\text{and}}^{(l), m}+ \text{\textit{forget}}_{\text{and}}^{(l), m},
\\
\textit{\text{all}}_\text{and}^{(L), m} =\text{\textit{overlap}}_{\text{and}}^{(l), m}+ \text{\textit{new}}_{\text{and}}^{(l), m}.
\end{aligned}\end{small}\end{equation}
Metrics for OR interactions {\small$\text{\textit{overlap}}_{\text{or}}^{(l), m}$}, {\small$\text{\textit{forget}}_{\text{or}}^{(l), m}$}, and {\small$\text{\textit{new}}_{\text{or}}^{(l), m}$} were defined in the similar way.

To evaluate the progress of learning target interactions and redundant interactions,
we also define the completeness of interactions encoded by each {\small$l$}-th layer~\textit{w.r.t.} all interactions {\small$\textit{\text{all}}_\text{and}^{(L), m}$}  encoded by the final layer, as
{\small $\textit{completeness}^{(l), m}_{\text{and}}=\textit{overlap}_{\text{and}}^{(l), m} /\ \textit{\text{all}}_\text{and}^{(L), m}$}. We define the redundancy of interactions in each {\small$l$}-th layer as the ratio of the interactions {\small$\text{\textit{forget}}_{\text{and}}^{(l), m}$} that are finally forgotten, as {\small $\textit{redundancy}^{(l), m}_{\text{and}}=\textit{forget}_{\text{and}}^{(l), m} /\ \textit{\text{all}}_\text{and}^{(l), m}$}.
What's more, we can define the metrics for OR interaction {\small $\textit{completeness}^{(l), m}_{\text{or}}$} and {\small $\textit{redundancy}^{(l), m}_{\text{or}}$} in the similar way.


\textbf{Results \& Analysis. } Fig.~\ref{fig:overlap}~(a) reports the average strength{\footnote{We normalized each AND interaction {\footnotesize$I_{\text{and}}(S\vert \boldsymbol{x},v^{(l)})$} extracted from the $l$-th layer of the target DNN as {\footnotesize$I_{\text{and}}(S\vert \boldsymbol{x},v^{(l)})\leftarrow I_{\text{and}}(S\vert \boldsymbol{x},v^{(l)})/\mathbb{E}_{\boldsymbol{x}}[\vert v^{(l)}(\boldsymbol{x}_{N})-v^{(l)}(\boldsymbol{x}_{\emptyset})\vert]$} for fair comparison. Each OR interaction was normalized in the similar way.}} of the overlapped AND interactions {\small$\text{\textit{overlap}}_{\text{and}}^{(l), m}$}, the forgotten AND  interactions {\small$\text{\textit{forget}}_{\text{and}}^{(l), m}$}, and newly emerged AND interactions {\small$\text{\textit{new}}_{\text{and}}^{(l), m}$}.
Fig.~\ref{fig:overlap}~(b) tracks the completeness and redundancy of the learned interactions through layers.
We discovered that even among different models on different tasks, most DNNs still tended to follow similar information-processing behaviors, as follows. 

$\bullet$\;
Fig.~\ref{fig:overlap}~(a) shows that {\small$\textit{\text{all}}_\text{and}^{(l), m}$} and {\small$\textit{\text{all}}_\text{and}^{(L), m}$} of low order have higher strength than those of high order, which indicates that DNNs usually encode stronger low-order (simple) interactions than high-order (complex) interactions.

$\bullet$\;
Fig.~\ref{fig:overlap}~(b) shows that {\small $\textit{completeness}^{(l), m}_{\text{and}}$} values of most interactions start increasing at early layers. Unlike low-order interactions, high-order interactions do not reach high completeness even in later layers. These indicate that the early and middle layers usually had already learned most target interactions that were finally used by DNNs. Morever, extremely high-order interactions are learned in later layers, but the learning is unstable.

$\bullet$\;
Fig.~\ref{fig:overlap}~(b) shows that {\small $\textit{redundancy}^{(l), m}_{\text{and}}$} values of most interactions first rise and then fall through layers, which indicates that the utility of later layers of DNNs was mainly to remove redundant interactions encoded by earlier layers.

$\bullet$\;
\textbf{Using our results to analyze the generalization power.} We further use the layerwise change of interactions on each specific DNN to analyze its generalization power. To this end, Section~\ref{sec:representation} will show a clear relationship between the order of interactions and the generalization power of interactions. In this way, Appendix~\ref{app_sec:discuss_distinc} introduces how to use the distribution of interactions to analyze the generalization power of features of different layers.

\subsection{Analyzing the representation capacity of a DNN}
\label{sec:representation}

Tracking salient interactions through layers also provides us a new perspective to understand how the representation capacity gradually changes during the forward propagation. 
It is because we find that the order (complexity) of interactions can well explain the generalization capacity and the instability of feature representations of a DNN.

$\bullet$\;
\textbf{Low-order interactions are more generalizable across models.}
According to Lemma~\ref{le:faithfulness}, we can disentangle the overall inference score based on the feature {\small$f^{(l)}(\boldsymbol{x})$} into the sum of effects of a few salient interactions, 
{\small$v^{(l)}(\boldsymbol{x}_T)
\approx v(\boldsymbol{x}_\emptyset)+
\sum\nolimits_{S \in \Omega^{(l)}_{\text{and}}: \emptyset\neq S\subseteq T }I_{\text{and}}(S\vert \boldsymbol{x_T},v^{(l)})+
\sum\nolimits_{S \in \Omega^{(l)}_{\text{or}}:S\cap T \ne \emptyset }I_{\text{or}}(S\vert \boldsymbol{x_T},v^{(l)}).$}
Thus, the generalization capacity of the feature {\small$f^{(l)}(\boldsymbol{x})$} can be explained by the generalization capacity of salient interactions.

\begin{figure*}[ht]
	\centering
	\includegraphics[width=\linewidth]{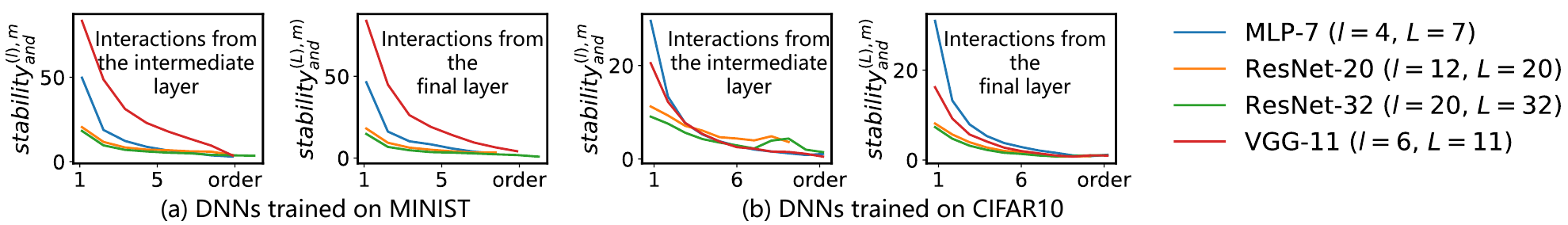}
	\vspace{-5pt}
	\caption{The relative stability ({$\textit{\text{stability}}^{(l),m}_{\text{and}}$}) of AND interactions decreased along with the order {\small$m$}. Low-order interactions were more stable to inevitable noises in data. See Appendix~\ref{app_sec:more_result} for results of OR interactions and Appendix~\ref{app_sec:exp_detail_layer} for the selected intermediate layer.}
	\label{fig:stability}
\end{figure*}

To this end, we consider that if multiple DNNs trained for the same task encode the same interaction, then this interaction is  regarded as well-generalized. 
Specifically, given two DNNs, {\small$v_A$} and {\small$v_B$}, trained for the same classification task and an input sample {\small$\boldsymbol{x}$}, 
we follow the settings in Section~\ref{sec:3.2.2} to extract two sets of $m$-order salient AND interactions from the $l_a$-th layer of the DNN {\small$v_A$} and the $l_b$-th layer of the DNN {\small$v_B$}, respectively, 
which are denoted by {\small$A_{\text{and}}^{(l_a),m}=\{S \subseteq N: \vert S\vert=m, \vert I_\text{and}(S \vert \boldsymbol{x}, v_A^{(l_a)} ) \vert > \tau{\footnotemark[7]} \}$} and {\small$B_{\text{and}}^{(l_b),m}$}.
Accordingly, let {\small$A_{\text{or}}^{(l_a),m}$} and {\small$B_{\text{or}}^{(l_b), m}$} represent sets of salient OR interactions of $m$-th order, respectively.
Then, we use the IoU metric to measure the generalization capacity of $m$-order interactions across different models.
\begin{equation}\begin{small}\begin{aligned}
\label{eq:sim}
IoU(A_{\text{and}}^{(l_a),m}, B_{\text{and}}^{(l_b),m})=
\frac{\vert A_{\text{and}}^{(l_a),m} \cap B_{\text{and}}^{(l_b),m} \vert}
{\vert A_{\text{and}}^{(l_a),m} \cup B_{\text{and}}^{(l_b),m} \vert}, \\
IoU(A_{\text{or}}^{(l_a),m}, B_{\text{or}}^{(l_b), m})=
\frac{\vert A_{\text{or}}^{(l_a),m} \cap B_{\text{or}}^{(l_b), m} \vert}
{\vert A_{\text{or}}^{(l_a),m} \cup B_{\text{or}}^{(l_b), m} \vert}.
\end{aligned}\end{small}\end{equation}

Large values of {\small$IoU(A_{\text{and}}^{(l_a),m}, B_{\text{and}}^{(l_b),m}) \text{ and } IoU(A_{\text{or}}^{(l_a),m},$} {\small $B_{\text{or}}^{(l_b), m})$}
mean that most $m$-order interactions encoded by a DNN can be well generalized to another DNN.

\textbf{\textit{Experiments.}}
Here, we examined the generalization capacity of interactions of different orders.
We used DNNs introduced in Section~\ref{sec:vis},~\textit{i.e.,}
MLP-7, VGG-11, ResNet-20, and ResNet-32~\citep{he2016deep} trained on the CIFAR-10 dataset for image classification, 
and DistilBERT, $\text{BERT}_\text{BASE}$, and XLNet~\citep{yang2019xlnet} fine-tuned on the SST-2 dataset for binary sentiment classification.

Fig.~\ref{fig:generalization} reports the average IoU value of AND interactions extracted from two DNNs over different input samples, {\small$\mathbb{E}_{\boldsymbol{x}}[IoU(A_{\text{and}}^{(l_a),m}, B_{\text{and}}^{(l_b),m})]$}, given each pair of DNNs trained for the same task.
We discovered low-order interactions extracted from different DNNs usually exhibited higher IoU values,~\textit{i.e.,} different DNNs trained for the same task usually encoded similar sets of salient low-order interactions.
This demonstrated low-order interactions could be better generalized across DNNs.
Notably, for each $m<n/2$, there are the same number $\tbinom{n}{m}$ of potential combinations for both $m\text{-order}$ interactions and $(n-m)\text{-order}$ interactions. Low-order ($m\text{-order}$) interactions are more generalizable than high-order interactions.

$\bullet$\;
\textbf{Low-order interactions are more stable to small noises.}
We discover that the order of interactions can also be used to explain the instability of feature representations of a DNN.  
According to Lemma~\ref{le:faithfulness}, the overall inference score based on the feature {\small$f^{(l)}(\boldsymbol{x})$} can be disentangled into the sum of the effects of a few salient AND-OR interactions.
Thus, the instability of the feature {\small$f^{(l)}(\boldsymbol{x})$} can be explained by the instability of salient interactions.

To this end, let us add a small Gaussian perturbation {\small $\boldsymbol{\epsilon} \sim \mathcal{N}(\boldsymbol{0}, \sigma^2\boldsymbol{I})$} to the input sample {\small$\boldsymbol{x}$}, in order to mimic inevitable noises/variations in data.
Although there may exist other noises in data, we just use Gaussian perturbation to represent noises/variations in data, which may still provide insights into real-world applications.
Thus, we use the following metrics to measure the relative stability of AND-OR interactions of each order {\small$m$}.
\begin{equation}\begin{small}\begin{aligned}
\textit{\text{stability}}^{(l),m}_{\text{and}} \!=\!{\mathbb{E}}_{\boldsymbol{x}}\!\!\!\mathop{\mathbb{E}}_{S\in\Omega_{\text{and}}^{(l),m}} \!\!\left[ \vert E^{(l)}_{\text{and}}(S,\boldsymbol{x})\vert \bigg/ \sqrt{\textit{Var}^{(l)}_{\text{and}}(S,\boldsymbol{x})}\  \right],\\
\textit{\text{stability}}^{(l),m}_{\text{or}} \!=\! {\mathbb{E}}_{\boldsymbol{x}}\!\!\!\mathop{\mathbb{E}}_{S\in\Omega_{\text{or}}^{(l),m}} \!\!\left[ \vert E^{(l)}_{\text{or}}(S,\boldsymbol{x})\vert \bigg/ \sqrt{\textit{Var}^{(l)}_{\text{or}}(S,\boldsymbol{x})}\  \right].
\end{aligned}\end{small}\end{equation}

where {\small$ E^{(l)}_{\text{and}}(S,\boldsymbol{x})=\mathbb{E}_{\boldsymbol{\epsilon}}[I_\text{and}(S \vert \boldsymbol{x} + \boldsymbol{\epsilon}, v^{(l)})]$} and 
{\small$\textit{Var}^{(l)}_{\text{and}}(S,\boldsymbol{x})= \textit{Var}_{\boldsymbol{\epsilon}}[I_\text{and}(S \vert \boldsymbol{x} + \boldsymbol{\epsilon}, v^{(l)})]$} denote the mean and variance of the AND interaction {\small$I_\text{and}(S \vert \boldsymbol{x} + \boldsymbol{\epsilon}, v^{(l)})$}~\textit{w.r.t.} Gaussian perturbations {\small$\boldsymbol{\epsilon}$}, which are encoded by the $l$-th layer of the DNN.
Similarly, {\small$E^{(l)}_{\text{or}}(S,\boldsymbol{x})$} and {\small$Var^{(l)}_{\text{or}}(S,\boldsymbol{x})$} represent the mean and variance of the OR interaction {\small$I_\text{or}(S \vert \boldsymbol{x} + \boldsymbol{\epsilon}, v^{(l)})$}~\textit{w.r.t.} noises {\small$\boldsymbol{\epsilon}$}.
Large values of {\small$\textit{\text{stability}}^{(l),m}_{\text{and}}$} and {\small$\textit{\text{stability}}^{(l),m}_{\text{or}}$} indicates that $m$-order interactions are stable to inevitable noises.

\textbf{\textit{Towards normalization of the noise.}}
As a common understanding, people usually think that higher-order interactions  contain more noise than low-order interactions, because they involve more input variables. However, it is noteworthy  that high-order interactions also obtain more input signals, and the signal-to-noise ratio of each interaction is relatively consistent over interactions of different orders. Thus {\small$\textit{\text{stability}}^{(l),m}_{\text{and}}$} and {\small$\textit{\text{stability}}^{(l),m}_{\text{or}}$} ensures a fair comparison of interactions of different orders {\small $m$}. Please see Appendix~\ref{app_sec:Signal_to_noise} for details.

\textbf{\textit{Experiments.}}
We conducted experiments to check the instability of AND-OR interactions of each order.
To this end, we added Gaussian perturbation {\small$\boldsymbol{\epsilon} \sim \mathcal{N}(\boldsymbol{0}, 0.02^2\boldsymbol{I})$} to each training sample.
Then, for each order {\small$m$}, we computed metrics {$\textit{\text{stability}}^{(l),m}_{\text{and}}$} based on DNNs, and the DNNs for testing have been introduced in Section~\ref{sec:vis}.
Fig.~\ref{fig:stability} shows that the relative stability {$\textit{\text{stability}}^{(l),m}_{\text{and}}$}  decreased along with the order {\small$m$}, which indicated that low-order interactions were more stable to inevitable noises in data than high-order interactions.
In other words, low-order interactions usually exhibited consistent effects {\small$I_\text{and}(S \vert \boldsymbol{x} + \boldsymbol{\epsilon}, v^{(l)})$} on the network output/intermediate-layer feature~\textit{w.r.t.} different noises {\small$\boldsymbol{\epsilon}$} than high-order interactions.
This indicated that low-order interactions were more likely to be generalized to similar samples (\textit{e.g.,} samples with small intra-class variations).

Thus, according to Figs.~\ref{fig:overlap},~\ref{fig:generalization},~\ref{fig:stability},
we discovered that for ResNet-20 trained on both the MNIST dataset and the CIFAR-10 dataset, their later layers usually exclusively forgot redundant high-order interactions without encoding new interactions, which were non-generalizable and unstable.
Besides, later layers of DistilBERT and $\text{BERT}_\text{BASE}$ trained on the SST-2 dataset usually forgot redundant  and non-generalizable high-order interactions.


\section{Conclusion, Discussion and Future Challenges}

In this paper, we use interaction primitives to represent knowledge encoded by the DNN.
The sparsity and the universal-matching property of interactions ensure the trustworthiness of taking interactions as symbolic primitive inference patterns encoded by a DNN.
Thus, we further quantify and track the newly emerged interaction primitives and the forgotten interaction primitives in each layer during the forward propagation, which provides new insights into the learning behavior of DNNs.
The layer-wise change of interactions potentially reveals the change of the generalization capacity and instability of feature representations of a DNN.

Alhtough the theory system of interaction-based explanation has been proposed for years, there are still many future challenges:
\\
1. using interaction primitives to represent the complex learning dynamics of a DNN;
\\
2. identifying and boosting the reliable/generalizable interaction primitives;
\\
3. aligning a DNN's detailed inference logic (interaction primitives) with human cognition.

\section*{Acknowledgements}
This work is partially supported by the National Science and Technology Major Project (2021ZD0111602), the National Nature Science Foundation of China (92370115, 62276165).
This work is also partially supported by Huawei Technologies Inc.

\section*{Impact Statement}

This paper uses interaction primitives to represent knowledge encoded by the DNN, and further quantify and track the layer-wise change of interaction primitives in the DNN during the forward propagation. The layer-wise change of interactions also reveals the change of the generalization capacity and instability of feature representations of the DNN. There are no ethical issues with this paper, and there are no potential societal consequences of this paper that need to be specifically highlighted.


\bibliography{example_paper}
\bibliographystyle{icml2024}

\newpage
\appendix
\onecolumn

\section{Detailed Analysis for Previous Studies Using Knowledge to Explain DNNs}
\label{app_sec:related}

Explaining and quantifying the exact knowledge encoded by a DNN presents a significant challenge to explainable AI.
So far, there has not existed a widely accepted definition of knowledge that enables us to accurately disentangle and quantify knowledge encoded by intermediate layers of a DNN, because it covers various aspects of cognitive science, neuroscience, and mathematics.
To this end, previous works have employed different methods to quantify knowledge encoded by a DNN.
Then, let us revisit previous studies from the perspective of three challenges mentioned in Section~1.

First,~\citet{bau2017network,kim2018interpretability} associated neurons with manually annotated semantics/concepts (knowledge).
However, these works could not quantify the exact amount of knowledge in the DNN, or discover new concepts emerged in intermediate layers. 
Second, learning interpretable neural networks with meaningful features in intermediate layers was another classic direction in explainable AI~\citep{zhang2020interpretable,ijcai2021p409,chen2019looks}.
Although these studies automatically learned meaningful concepts without human annotations, they did not provide a mathematically guaranteed boundary for each concept/knowledge. 
Thus, these works could not quantify the exact amount of newly emerged/forgotten/unexplainable knowledge in each layer.

Third,
the information-bottleneck theory~\citep{shwartz2017opening,michael2018on} used the mutual information between inputs and intermediate-layer features to quantify knowledge encoded by the DNN.
However, the mutual information could only measure the overall information contained in each feature, but could not accurately quantify exact knowledge represented by the newly emerged information and the forgotten information.
Besides,~\citet{kolchinsky2018caveats} showed the mutual information was difficult to measure accurately, and
\citet{wang2022pacbayes,michael2018on} discovered the mutual information had mathematical flaws in explaining the generalization power of a DNN.

Fourth, 
\citet{Liang2020Knowledge} disentangled feature components from each layer, which could be reconstructed by features in other layers, so as to evaluate the changes of features in different layers. 
However, the changes of features in different layers could not be aligned to the same feature space for fair comparison, and could not be employed to explain the generalization capacity of the DNN.

\section{Comparison between Interaction-based Explanation and Attribution Interpretability Methods}
\label{app_sec:comp}

We compare our interaction-based explanation with attribution interpretability methods from the following three perspectives. 

\begin{figure*}[h!]
	\centering
	\includegraphics[width=\linewidth]{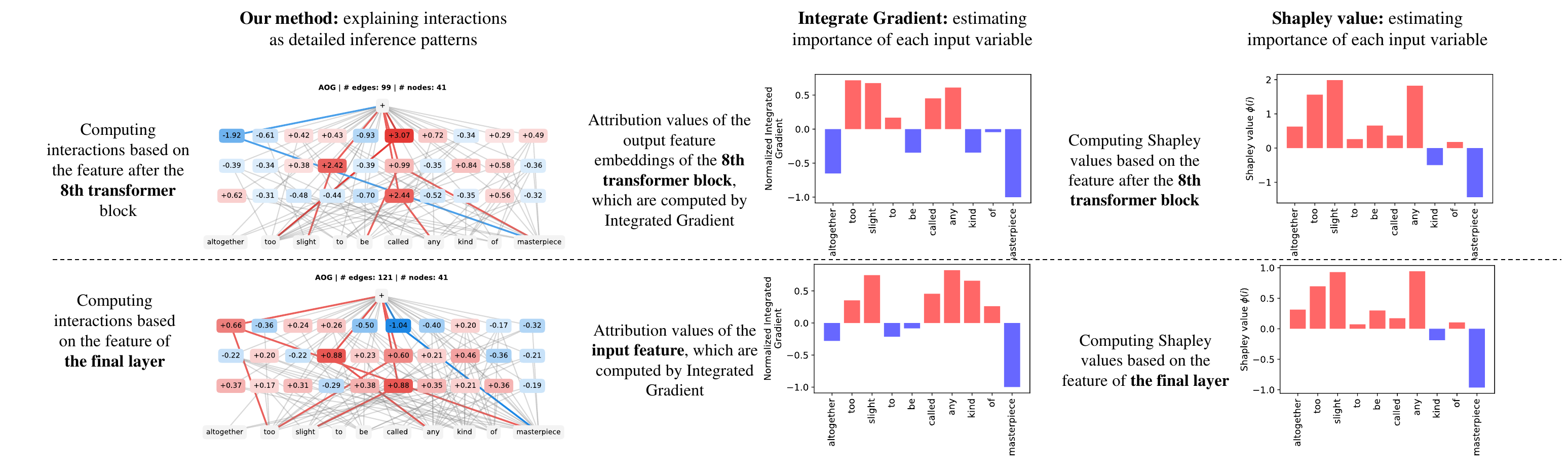}
	\caption{The comparison of explanation results between our interaction-based method, Integrated Gradient, and Shapley value.}
    \label{fig:comp}
\end{figure*}

$\bullet$ 
\textbf{Perspective 1: Whether the method can explain the detailed inference logic of a DNN.}
 We have conducted a new experiment to show explanation results of other interpretability methods for comparison. 
 Figure~\ref{fig:comp} shows that the main difference between our interaction-based explanation and traditional attribution methods (such as Integrated Gradient~\cite{sundararajan2017axiomatic}, Shapley value~\cite{Shapley1953}, and \textit{etc}.) is that the interaction-based explanation precisely shows the detailed inference logic of the DNN, while attribution methods can only provide the importance score of each input variable to the network output. 
Thus, our method can provide a more precise explanation than attribution methods, and the faithfulness of our method is theoretically ensured by the universal matching property in Theorem 3.3.
 
$\bullet$ 
\textbf{Perspective 2: Theoretical connections between AND-OR interactions and attribution explanation methods.}
 Besides, our AND-OR interactions can be considered as the elementary factors that determine the Shapley value~\cite{Shapley1953}. 
 The Shapley value is a widely accepted standard attribution method, which assigns the numerical attribution of each input variable to the network output score. 
 The Shapley value satisfies \textit{linearity, nullity, symmetry}, and \textit{efficiency} axioms. 
 The Shapley value $\phi(i)$ of each input variable  $i \in N$ can be rewritten as a re-allocation of AND-OR interactions, \textit{i.e.,},  
 $\phi(i) = \sum\nolimits_{S \subseteq N, S \ni i} \frac{1}{\vert S \vert} I_{\text{and}}(S) + \sum\nolimits_{S \subseteq N, S \ni i} \frac{1}{\vert S \vert} I_{\text{or}}(S)$. 
 It means that the computation of the Shapley value $\phi(i)$ can be explained as uniformly allocating the effect of each interaction $I(S)$ to its compositional input variables $i\in S$.

$\bullet$ 
\textbf{Perspective 3: If each feature dimension in an intermediate layer does not have a clear receptive field, the explanation based on this feature dimension's attribution will not have clear semantic meaning. In this case, we can only use interactions to explain the DNN.}
For example, in an MLP, each feature dimension in an intermediate layer does not have a clear receptive field on input variables,  
\textit{i.e.,}, this feature dimension does not have a clear physical meaning. In this case, visualizing the attribution of the feature dimension does not provide an intuitive explanation. so that it is meaningless to compute the attribution/importance of these features. 
In comparison, Figure.~\ref{fig:comp} shows that interactions extracted from a high layer still have clear physical meaning, and interactions used for computing a high-layer feature can also be aligned with interactions for computing a low-layer feature.
\section{Proving the OR Interaction Can Be Considered A Specific AND Interaction} 
\label{app_sec:or}

The OR interaction {\small$I_{\text{or}}(S \vert \boldsymbol{x})$} can be considered as a specific AND interaction interaction {\small$I_{\text{and}}(S \vert \boldsymbol{x})$}, when we we inverse the definition of masked states and unmasked states of the input variable.

Specifically, given an input sample {\small$\boldsymbol{x}\in \mathbb{R}^n$}, 
let {\small$\boldsymbol{x}_{N \setminus T}$} denote the masked sample obtained by masking input variables in {\small $T$}, while leaving variables in {\small $N \setminus T$} unaltered.
 Here, we mask the input variable {\small$i\in T$} to the baseline value {\small$b_i$} to represent its masked state, as follows.
\begin{equation}\begin{small}\begin{aligned}
\label{eq:masked_state}
    (\boldsymbol{x}_{N \setminus T})_i =\begin{cases}
x_i,& \text{$i\in N \setminus T$}\\
b_i,& \text{$i\in T$}
\end{cases}
\end{aligned}\end{small}\end{equation}

Then, let us consider the masked sample {\small$\boldsymbol{x}^{\prime}_T$}, where we inverse the definition of the masked state and the unmasked state of each input variable to obtain this masked sample.
That is, we mask input variables in the set {\small $N \setminus T$} to baseline values, and keep variables in {\small $T$} unchanged, as follows.
\begin{equation}\begin{small}\begin{aligned}
\label{eq:inversed_masked_state}
    (\boldsymbol{x}^{\prime}_T)_i =\begin{cases}
x_i,& \text{$i\in T$}\\
b_i,& \text{$i\in N \setminus T$}
\end{cases}
\end{aligned}\end{small}\end{equation}

Thus, the OR interaction {\small$I_{\text{and}}(S \vert \boldsymbol{x})$} in Eq.~2 in main paper can be represented by the specific AND interaction {\small$I_{\text{and}}(S \vert \boldsymbol{x}^{\prime})$}, as follows.
\begin{equation}\begin{small}\begin{aligned}
\label{eq:relationship_and_or_interactions}
  I_{\text{or}}(S \vert \boldsymbol{x}) 
  &= - \sum\nolimits_{T \subseteq S} (-1)^{\vert S \vert - \vert T \vert} v(\boldsymbol{x}_{N \setminus T}),
  \\
    &=- \sum\nolimits_{T \subseteq S} (-1)^{\vert S \vert - \vert T \vert} v(\boldsymbol{x}^{\prime}_T), 
    \\
    &= - I_{\text{and}}(S \vert \boldsymbol{x}^{\prime}).
\end{aligned}\end{small}\end{equation}

In this way, based on Eq.~(\ref{eq:relationship_and_or_interactions}), the proven sparsity of AND interactions in~\citep{ren2023we} also proves the 
sparsity of OR interactions,~\textit{i.e.,}
most well-trained DNNs usually encode a small number of OR interactions.

\section{Discussion on Techniques and Limitations of Classifier Probe}
\label{app_sec:discuss_probe}
Probing classifiers have become one of the prominent methodologies  for interpreting and analyzing deep neural network models. This approach involves training a classifier to predict a specific linguistic property based on the representations generated by a model. In our study, we utilize this technique by selecting features from the mid-layers of a neural network. These selected features are then used to train a linear classifier, enabling us to assess and understand the knowledge contained within the mid-layer features of the neural network. However, it's important to acknowledge that the probing classifiers approach is not without its drawbacks. ~\citet{belinkov2022probing} have outlined the limitations of this framework, and we will explore these challenges in the following discussion.

$\bullet$\; 
A primary challenge arises in how we interpret the performance from the probing classifier, particularly in selecting an appropriate baseline for comparison. Our study diverges from the traditional method of directly comparing overall model performance. Instead, we concentrate on the knowledge contained within the middle layer features, specifically their direct applicability to classification tasks. In fact, our findings reveal a high degree of similarity in the knowledge interpreted across adjacent layers, which partly reflects the faithfulness of our work.

$\bullet$\; 
The second challenge concerns the selection of the classifier's structure. \citet{pimentel2020information} contend that to obtain the most accurate estimate of the information a model possesses about a given property, it is advisable to use the most complex probe available. However, our research is not focused on the ultimate classifiability of the intermediate layer. Instead, we are interested in determining the extent to which the features of this layer can be directly applied to the classification task. In light of this, we opt for one of the simplest classifier structures available: the linear classifier. This choice is driven by our specific objective of evaluating the direct applicability of middle layer features, rather than maximizing the classification potential of the probe.

$\bullet$\; 
The third challenge addresses the disconnect between the probing classifier {\small $g$} and the original model {\small $f$}. This implies that the knowledge inferred from the classifier may not always align with what is actually utilized by the original model. This is a good question. In fact, this perspective is central to our research. Our findings indicate that neural networks tend to learn numerous redundant features in the middle layers, which are subsequently forgotten in the later layers. Our methodology offers a quantifiable analysis of the variation in knowledge across different layers, shedding light on how information is processed and transformed within the network. 

$\bullet$\;  
The fourth challenge concerns the imperfection of the dataset used for training. Specifically, the classifier is unable to exhaustively uncover all the knowledge present due to the limitations inherent in the dataset. This is a real drawback, as it is not feasible to use an all-encompassing dataset for perfect training. Our approach mitigates this issue by training the classifier on the same dataset as the original model. This strategy aims to ensure as fair and balanced a training process as possible, while acknowledging the constraints of the dataset while striving.

In summary, while employing probing classifiers in the interpretation of neural network models does introduce specific challenges, our approach aims to maximize the potential and ensure a thorough and insightful analysis of neural network models.




\section{Discussion on the Bias Introduced by Masking Input Variables}
\label{app_sec:discuss_baseline_bias}
In attribution method research, a prevalent approach involves utilizing a designated baseline value to obscure input variables in a DNN\citep{lundberg2017unified, ancona2019explaining, fong2019understanding}. This technique measures the impact of these masked inputs on the network's output, thereby estimating the significance of each input variable. Nevertheless, research by ~\citet{jain2022missingness} indicates that this current method of masking may introduce substantial bias in the model's predictions. Specifically, it has been observed that the DNN tends to make errors influenced more by the areas subjected to masking than by the unmasked features.

In efforts to mitigate this bias, \citet{ren2023can} have proposed utilizing causal patterns to scrutinize the reliability of baseline values. More importantly, they have established that causal patterns can be interpreted as the fundamental logic behind the concept of the Shapley value. Building upon this, they have proposed a novel methodology for determining optimal baseline values. The efficacy of this approach is underscored by the positive outcomes observed in various experimental settings.

\newpage
\section{Proof of Theorem~3.3} 
\label{app_sec:proof_them1}
\textbf{Theorem 3.3}
\textit{
Given an input sample {\small $\boldsymbol{x}\in\mathbb{R}^{n}$}, the network output score {\small$v(\boldsymbol{x}_T)$} on each masked input samples {\small$\{\boldsymbol{x}_T\vert T\subseteq N\}$} can be decomposed into effects of AND interactions and OR interactions,
subject to {\small$I_{\text{and}}(\emptyset \vert \boldsymbol{x}) = v_\text{and}(\boldsymbol{x}_\emptyset)=v(\boldsymbol{x}_\emptyset)$} and {\small$I_{\text{or}}(\emptyset \vert \boldsymbol{x}) = v_\text{or}(\boldsymbol{x}_\emptyset)=0$}.
\begin{equation}\begin{small}\begin{aligned}
\label{eq:harsanyi_interaction_sum}
v(\boldsymbol{x}_T)
&= v_{\text{and}}(\boldsymbol{x}_T)+v_{\text{or}}(\boldsymbol{x}_T) \\
&= \sum\nolimits_{S\subseteq T} I_{\text{and}}(S|\boldsymbol{x}_T) + \sum\nolimits_{S\cap T \ne \emptyset} I_{\text{or}}(S|\boldsymbol{x}_T).
\end{aligned}\end{small}\end{equation}}

\begin{proof}
Let us first focus on the sum of AND interactions, as follows.
\begin{equation}\begin{small}\begin{aligned}
    \sum\nolimits_{S \subseteq T} I_{\text{and}}(S\vert \boldsymbol{x_T}) 
    &=  \sum\nolimits_{S \subseteq T} \sum\nolimits_{L \subseteq S} (-1)^{\vert S \vert - \vert L \vert} v_{\text{and}}(\boldsymbol{x}_L) \\
    &= \sum\nolimits_{L \subseteq T} \sum\nolimits_{S: L \subseteq S \subseteq T} (-1)^{\vert S \vert - \vert L \vert} v_{\text{and}}(\boldsymbol{x}_L) \\
    &= \underbrace{v_{\text{and}}(\boldsymbol{x}_T)}_{L = T} + \sum\nolimits_{L \subseteq T, L \neq T} v_{\text{and}}(\boldsymbol{x}_L) \cdot \underbrace{\sum\nolimits_{m=0}^{\vert T \vert - \vert L \vert} (-1)^m}_{=0} \\
    &= v_{\text{and}}(\boldsymbol{x}_T).
\end{aligned}\end{small}\end{equation}

Then, let us concentrate on the the sum of OR interactions, as follows.
\begin{equation}\begin{small}\begin{aligned}
\label{eq:prove_th1_1}
        \sum\nolimits_{S \cap T \neq \emptyset} I_{\text{or}}(S\vert \boldsymbol{x_T})
        &= - \sum\nolimits_{S \cap T \neq \emptyset, S \neq \emptyset} \sum\nolimits_{L \subseteq S} (-1)^{\vert S \vert - \vert L \vert} v_{\text{or}}(\boldsymbol{x}_{N \setminus L}) \\
        &= - \sum\nolimits_{L \subseteq N} \sum\nolimits_{S: S \cap T \neq \emptyset, S \supseteq L} (-1)^{\vert S \vert - \vert L \vert} v_{\text{or}}(\boldsymbol{x}_{N \setminus L}) \\
        &= - \underbrace{v_{\text{or}}(\boldsymbol{x}_{\emptyset})}_{L = N} - \underbrace{v_{\text{or}}(\boldsymbol{x}_T)}_{L = N \setminus T} \cdot \underbrace{\sum_{\vert S_2 \vert = 1}^{\vert T \vert} C_{\vert T \vert}^{\vert S_2 \vert} (-1)^{\vert S_2 \vert}}_{=-1} \\
        &\quad- \sum_{L \cap T \neq \emptyset, L \neq N} v_{\text{or}}(\boldsymbol{x}_{N \setminus L}) \cdot \sum_{S_1 \subseteq N\setminus T \setminus L} \underbrace{\sum_{\vert S_2 \vert = \vert T \cap L \vert}^{\vert T \vert} C_{\vert T \vert - \vert T \cap L \vert}^{\vert S_2 \vert - \vert T \cap L \vert} (-1)^{\vert S_1 \vert + \vert S_2 \vert}}_{=0} \\
        &\quad- \sum_{L \cap T = \emptyset, L \neq N \setminus T} v_{\text{or}}(\boldsymbol{x}_{N \setminus L}) \cdot \underbrace{\sum_{S_2 \subsetneqq T} \sum_{S_1 \subseteq N\setminus T \setminus L} (-1)^{\vert S_1 \vert + \vert S_2 \vert}}_{=0} \\
        &= v_{\text{or}}(\boldsymbol{x}_T) - v_{\text{or}}(\boldsymbol{x}_{\emptyset})
\end{aligned}\end{small}\end{equation}.

Thus, we obtain {\small$v_{\text{or}}(\boldsymbol{x}_T) = \sum\nolimits_{S \cap T \neq \emptyset} I_{\text{or}}(S) + v_{\text{or}}(\boldsymbol{x}_{\emptyset})$}, according to Eq.~(\ref{eq:prove_th1_1}).
Thus, the output score {\small$v(\boldsymbol{x}_T)$} of the DNN on the masked sample {\small$\boldsymbol{x}_T$} can be represented as the sum of effects of AND-OR interactions.
\begin{equation}\begin{small}\begin{aligned}
\label{eq:prove_th1_2}
        v(\boldsymbol{x}_T) 
        &= v_{\text{and}}(\boldsymbol{x}_T) + v_{\text{or}}(\boldsymbol{x}_T) \\
        &= \sum\nolimits_{S \subseteq T} I_{\text{and}}(S\vert \boldsymbol{x_T}) + \sum\nolimits_{S \cap T \neq \emptyset, S \neq \emptyset} I_{\text{or}}(S\vert \boldsymbol{x_T}) + v_{\text{or}}(\boldsymbol{x}_{\emptyset}) \\
        &= \sum\nolimits_{S \subseteq T, S \neq \emptyset} I_{\text{and}}(S\vert \boldsymbol{x_T}) + v_{\text{and}}(\boldsymbol{x}_{\emptyset}) + \sum\nolimits_{S \cap T \neq \emptyset} I_{\text{or}}(S\vert \boldsymbol{x_T}) + v_{\text{or}}(\boldsymbol{x}_{\emptyset}) \\
        &= v(\boldsymbol{x}_{\emptyset}) + \sum\nolimits_{S \subseteq T, S \neq \emptyset} I_{\text{and}}(S\vert \boldsymbol{x_T}) + \sum\nolimits_{S \cap T \neq \emptyset} I_{\text{or}}(S\vert \boldsymbol{x_T})\\
        &= \sum\nolimits_{S\subseteq T} I_{\text{and}}(S|\boldsymbol{x}_T) + \sum\nolimits_{S\cap T \ne \emptyset} I_{\text{or}}(S|\boldsymbol{x}_T)
        .
\end{aligned}\end{small}\end{equation}

Thus, Theorem~3.3 is proven.

\end{proof}

\newpage
\section{Proof of Lemma~3.4} 
\label{app_sec:proof_lemma1}

\textbf{Lemma 3.4} (\textbf{Proving interactions as primitive inference patterns})
\textit{
Given an input sample {\small $\boldsymbol{x}\in\mathbb{R}^{n}$}, the network output on all {\small $2^n$} masked input samples {\small$\{\boldsymbol{x}_S\vert S\subseteq N\}$} can be universally matched by a small set of salient interactions.
\begin{equation}\begin{small}\begin{aligned}
\label{{le:faithfulness}}
v(\boldsymbol{x}_T)
&= v_{\text{and}}(\boldsymbol{x}_T)+v_{\text{or}}(\boldsymbol{x}_T)
=\sum\limits_{ S\subseteq T} I_{\text{and}}(S|\boldsymbol{x}_T) + \sum\limits_{S\cap T \ne \emptyset} I_{\text{or}}(S|\boldsymbol{x}_T)
\\
&\approx v(\boldsymbol{x}_\emptyset)+
\sum\nolimits_{S \in \Omega^{\text{and}}_{\text{salient}}:  \emptyset\ne S\subseteq T} I_{\text{and}}(S\vert \boldsymbol{x_T})+
\sum\nolimits_{S \in \Omega^{\text{or}}_{\text{salient}}: S\cap T \ne \emptyset}I_{\text{or}}(S\vert \boldsymbol{x_T}).
\end{aligned}\end{small}\end{equation}}

\begin{proof}

\citet{Ren_2023_CVPR} have proven that under some common conditions{\color{red}\footnotemark[1]},
the output {\small$v_{\text{and}}(\boldsymbol{x}_T)$} of a well-trained DNN on all {\small$2^n$} masked samples $\{\boldsymbol{x}_T|T\subseteq N\}$ can be universally approximated by a small number of AND interactions {\small$T \in \Omega^{\text{and}}_{\text{salient}}$} with salient effects {\small$I_{\text{and}}(T \vert \boldsymbol{x})$} on the network output, subject to {\small $\vert \Omega^{\text{and}}_{\text{salient}} \vert \ll 2^n$}.

Besides, as proven in Appendix~\ref{app_sec:or}, the OR interaction can be considered as a specific AND interaction.
Thus, the output {\small$v_{\text{or}}(\boldsymbol{x}_T)$} of a well-trained DNN on all {\small$2^n$} masked samples $\{\boldsymbol{x}_T|T\subseteq N\}$ can be universally approximated by a small number of OR interactions {\small$T \in \Omega^{\text{or}}_{\text{salient}}$} with salient effects {\small$I_{\text{or}}(T \vert \boldsymbol{x})$} on the network output, subject to {\small $\vert \Omega^{\text{or}}_{\text{salient}} \vert \ll 2^n$}.

In this way, Eq.~(\ref{eq:prove_th1_2}) can be further approximated as
\begin{equation}\begin{small}\begin{aligned}
\label{eq:prove_lemma1_2}
        v(\boldsymbol{x}_T) 
        &= v_{\text{and}}(\boldsymbol{x}_T) + v_{\text{or}}(\boldsymbol{x}_T)
         \\
        &= v(\boldsymbol{x}_{\emptyset}) + \sum\nolimits_{S \subseteq T} I_{\text{and}}(S\vert \boldsymbol{x_T}) + \sum\nolimits_{S \cap T \neq \emptyset} I_{\text{or}}(S\vert \boldsymbol{x_T})
        \\
        &\approx v(\boldsymbol{x}_\emptyset)+
\sum\nolimits_{S \in \Omega^{\text{and}}_{\text{salient}}:  \emptyset\ne S\subseteq T} I_{\text{and}}(S\vert \boldsymbol{x_T})+
\sum\nolimits_{S \in \Omega^{\text{or}}_{\text{salient}}: S\cap T \ne \emptyset}I_{\text{or}}(S\vert \boldsymbol{x_T}).
\end{aligned}\end{small}\end{equation}

Thus, Lemma~3.4 is proven.

\end{proof}

\newpage
\section{Table of Metrics Used in the Paper}
\label{app_sec:table}

For a better understanding of our paper, we conclude all used metrics into the following table, where we clarify the formulation and the physical meaning of each metric.

\begin{figure*}[ht]
	\centering
	\includegraphics[width=\linewidth]{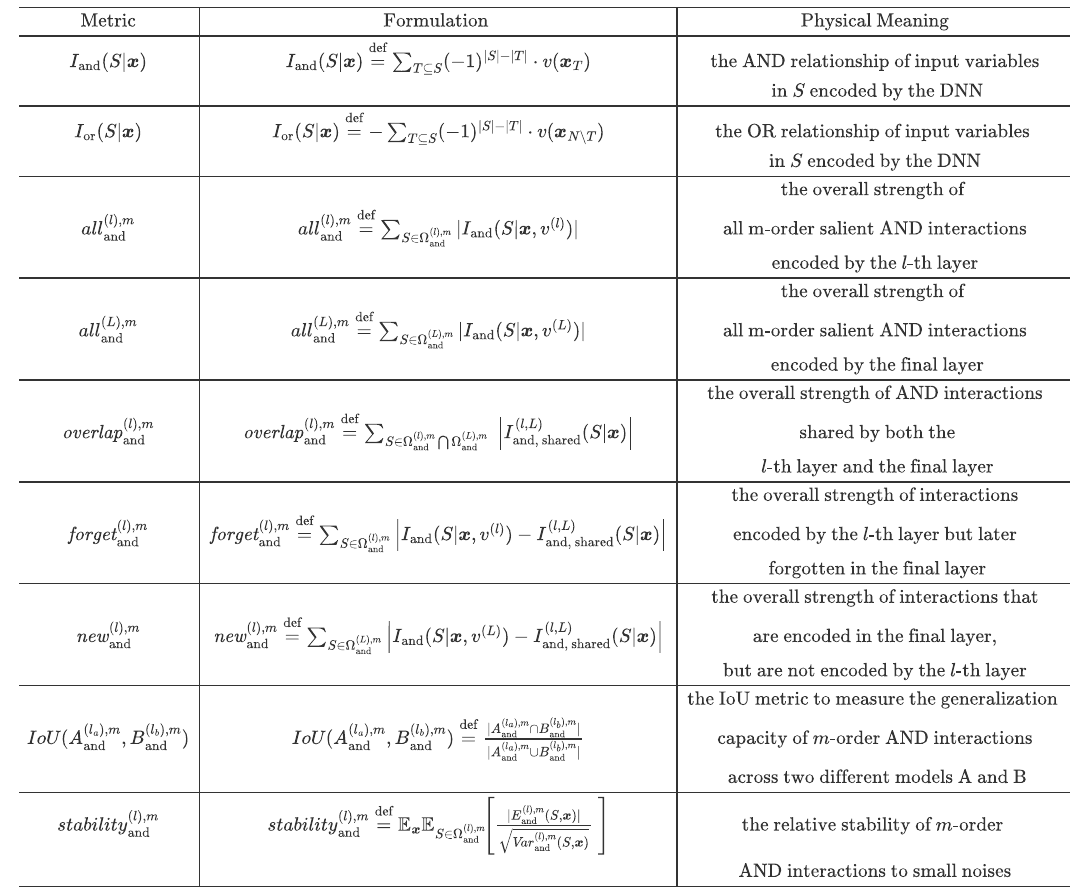}
\end{figure*}

\newpage
\section{More Experimental Results}
\label{app_sec:more_result}
\subsection{Experimental Results of OR Interactions}

\begin{figure*}[ht]
	\centering
	\includegraphics[width=\linewidth]{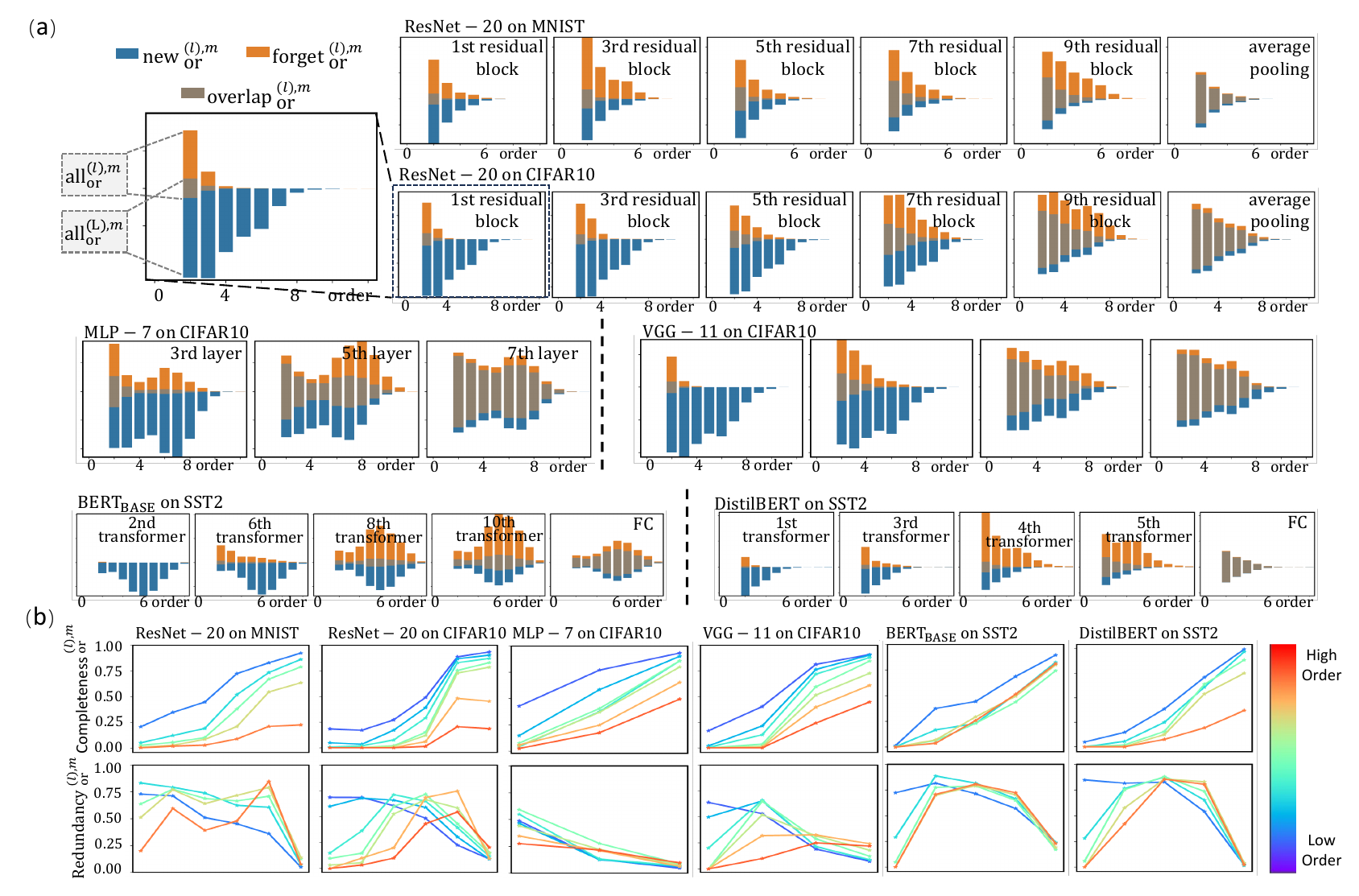}
	\vspace{-8pt}
	\caption{(a)~Tracking the change of the average strength of the overlapped {\small$\text{\textit{overlap}}_{\text{or}}^{(l), m}$}, forgotten {\small$\text{\textit{forget}}_{\text{or}}^{(l), m}$}, and newly emerged interactions {\small$\text{\textit{new}}_{\text{or}}^{(l), m}$} through different layers. For each subfigure, the total length of the orange bar and the grey bar equals to {\small$\textit{\text{all}}_\text{or}^{(l), m}$}, and the total length of the blue bar and the grey bar equals to {\small$\textit{\text{all}}_\text{or}^{(L), m}$}
	(b)~Tracking the change of {\small $\textit{completeness}^{(l), m}_{\text{or}}$} and {\small $\textit{redundancy}^{(l), m}_{\text{or}}$} through different layers. We do not show interactions of the highest four orders, because almost no interactions of extremely high orders were learned.}
	\label{fig:overlap_or}	
	\vspace{5pt}
\end{figure*}

\begin{figure*}[h!]
	\centering
	\includegraphics[width=\linewidth]{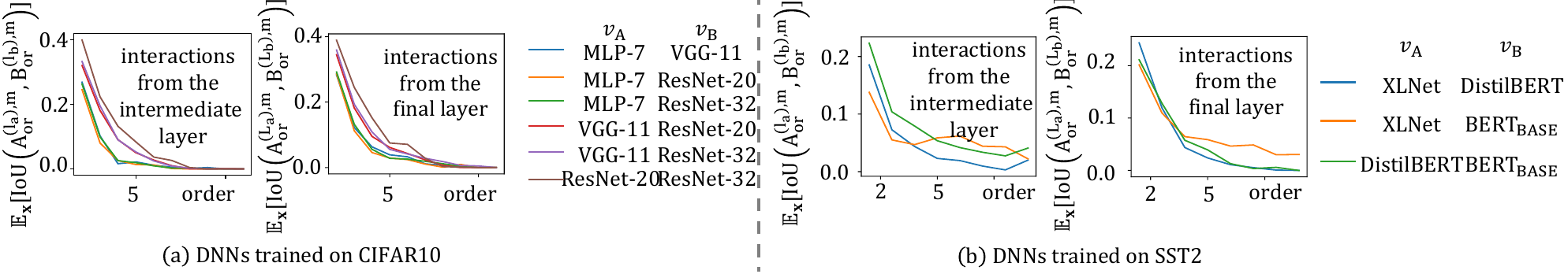}
	\vspace{-8pt}
	\caption{Average IoU values of OR interactions extracted from two DNNs trained for the same task over different input samples. Low-order interactions usually exhibited higher IoU values, which indicated that low-order interactions could be better generalized across DNNs than high-order interactions. Appendix~\ref{app_sec:exp_detail_layer} introduces the selected intermediate layer for each DNN.}
	\label{fig:generalization_or}
	\vspace{9pt}
\end{figure*}

\begin{figure*}[h!]
	\centering
	\includegraphics[width=\linewidth]{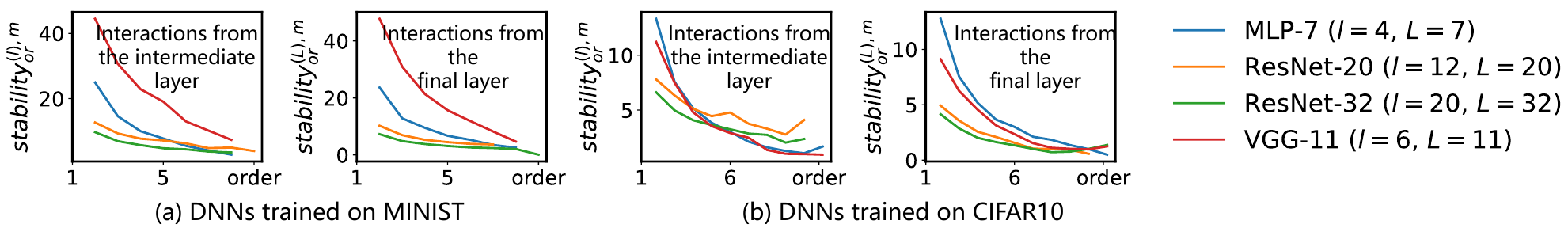}
	\vspace{-8pt}
	\caption{The relative stability {$\textit{\text{stability}}^{(l),m}_{\text{or}}$} of OR interactions decreased along with the order {\small$m$}. It indicated that low-order interactions were more stable to inevitable noises in data. Appendix~\ref{app_sec:exp_detail_layer} introduces the selected intermediate layer for each DNN.}
	\label{fig:stability_or}
\end{figure*}

\subsection{Experimental Results on Tabular Datasets}
We also trained the MLP-7 model on two tabular datasets (the UCI census dataset and the commercial dataset), and tracked the layer-wise change of interactions during the forward propagation in the MLP-7 model. 
Specifically, we calculated metrics $\text{\textit{overlap}}_{\text{and}}^{(l), m}$, $\text{\textit{forget}}_{\text{and}}^{(l), m}$, and $\text{\textit{new}}_{\text{and}}^{(l), m}$ to quantify the overlapped AND interactions, forgotten AND interactions, and newly emerged AND interactions, respectively.
We also calculated the $\text{completeness}_{\text{and}}^{(l),m}$ metric and the $\text{redundance}_{\text{and}}^{(l),m}$ metric to evaluate the progress of learning target AND interactions and removing redundant AND interactions. 

\begin{figure*}[h!]
	\centering
	\includegraphics[width=\linewidth]{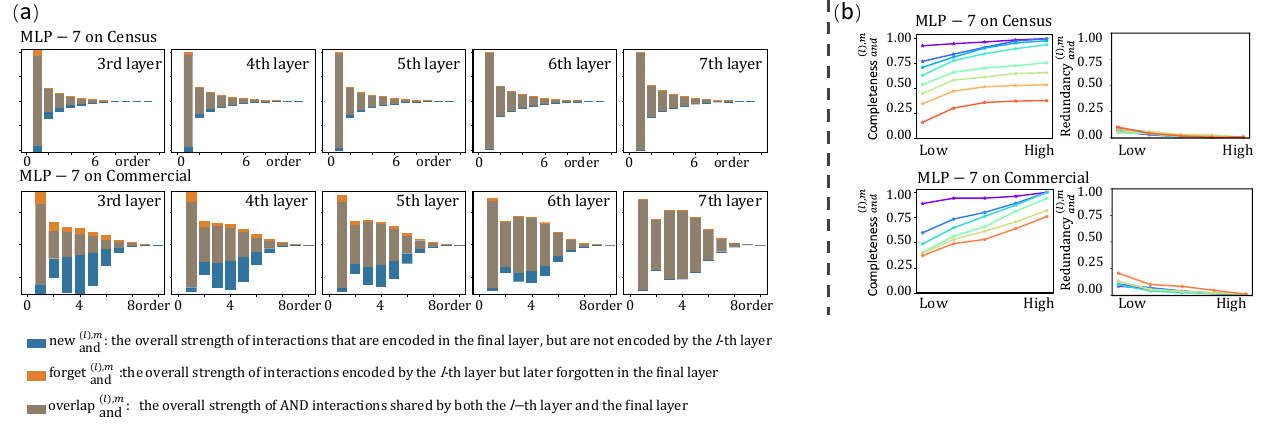}
        \vspace{-8pt}
	\caption{
	(a)~Tracking the change of the average strength of the overlapped {\small$\text{\textit{overlap}}_{\text{or}}^{(l), m}$}, forgotten {\small$\text{\textit{forget}}_{\text{or}}^{(l), m}$}, and newly emerged interactions {\small$\text{\textit{new}}_{\text{or}}^{(l), m}$} through different layers. For each subfigure, the total length of the orange bar and the grey bar equals to {\small$\textit{\text{all}}_\text{or}^{(l), m}$}, and the total length of the blue bar and the grey bar equals to {\small$\textit{\text{all}}_\text{or}^{(L), m}$}
	(b)~Tracking the change of {\small $\textit{completeness}^{(l), m}_{\text{or}}$} and {\small $\textit{redundancy}^{(l), m}_{\text{or}}$} through different layers. We do not show interactions of the highest four orders, because almost no interactions of extremely high orders were learned.}
	\label{fig:tabular}	
	\vspace{10pt}
\end{figure*}

Figure~\ref{fig:tabular} (a) reports the average strength of the overlapped, forgotten, and newly emerged interactions through different layers, and
Figure~\ref{fig:tabular} (b) tracks the completeness and redundancy of the learned interactions through layers. 
It shows that (1) DNNs encode stronger low-order (simple) interactions than high-order (complex) interactions. (2) The early and middle layers usually had already learned most target interactions that were finally used by DNNs. Moreover, extremely high-order interactions are learned in later layers, but the learning is unstable. (3) DNNs quickly learn all target interactions without learning many redundant interactions. These conclusions are similar to the conclusions obtained on models trained on MNIST, CIFAR-10, SST-2  in the paper.

\newpage
\section{Ablation Study of {$\kappa$} Value in Section~\ref{sec:3.2.2}}
\label{app_sec:ablation}
We find that the small noises in the output can significantly change the interaction effect. To remove the tiny noise in the model output and then extract relatively clean interactions, we define the new model output score {\small$v(\boldsymbol{x})$}.
\begin{equation}\begin{small}\begin{aligned}
v(\boldsymbol{x}) &= \text{log} \frac{p(y=y^{\text{truth}} \vert \boldsymbol{x})}{1 - p(y=y^{\text{truth}} \vert \boldsymbol{x})}-\delta_{N} \\
v(\boldsymbol{x}_{T}) &= \text{log} \frac{p(y=y^{\text{truth}} \vert \boldsymbol{x}_{T})}{1 - p(y=y^{\text{truth}} \vert \boldsymbol{x}_{T})}-{\delta}_{T}
\label{app:pq}
\end{aligned}\end{small}\end{equation}
where {\small$\delta_T, s.t.\; \forall T\subseteq N, \vert \delta_T \vert < \kappa$} is a learnable residual proposed to model and remove the tiny noise in the output {\small$v^{(l)}(\boldsymbol{x}_{T})$}, so as to extract relatively clean interactions.
{\small$\delta_T$} is constrained to a small range {\small$\kappa=0.04\cdot\vert v^{(l)}(\boldsymbol{x}_{N})-v^{(l)}(\boldsymbol{x}_{\emptyset})\vert$}.

We conducted the ablation study to verify that the extraction of interactions is relatively robust to the {\small $\kappa$} value. Given a well-trained DNN (or linear classifier learned by intermediate layer's features) and an input sample {\small $\boldsymbol{x}\in\mathbb{R}^{n}$}, we verify that the interactions extracted are stable in different {\small $\kappa$} value settings. To this end, we used the ResNet-20 trained on the CIFAR-10 (introduced in Section 3.2.1) and extracted the all AND-OR interactions encoded in different setting, \textit{e.g.,} {\small$\kappa=0.03\cdot\vert v^{(l)}(\boldsymbol{x}_{N})-v^{(l)}(\boldsymbol{x}_{\emptyset})\vert$}, {\small$\kappa=0.04\cdot\vert v^{(l)}(\boldsymbol{x}_{N})-v^{(l)}(\boldsymbol{x}_{\emptyset})\vert$} and {\small$\kappa=0.05\cdot\vert v^{(l)}(\boldsymbol{x}_{N})-v^{(l)}(\boldsymbol{x}_{\emptyset})\vert$}.

\begin{figure*}[h!]
	\centering
	\includegraphics[width=\linewidth]{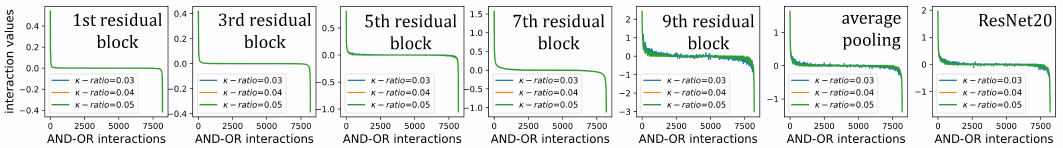}
	\caption{The extracted AND-OR interactions encoded in different setting, \textit{e.g.,} {\small$\kappa=0.03\cdot\vert v^{(l)}(\boldsymbol{x}_{N})-v^{(l)}(\boldsymbol{x}_{\emptyset})\vert$}, {\small$\kappa=0.04\cdot\vert v^{(l)}(\boldsymbol{x}_{N})-v^{(l)}(\boldsymbol{x}_{\emptyset})\vert$} and {\small$\kappa=0.05\cdot\vert v^{(l)}(\boldsymbol{x}_{N})-v^{(l)}(\boldsymbol{x}_{\emptyset})\vert$} through different layers, as well as the raw ResNet-20. We rearranged all of the AND-OR interactions in the order of the interaction value strength at {\small$\kappa=0.04\cdot\vert v^{(l)}(\boldsymbol{x}_{N})-v^{(l)}(\boldsymbol{x}_{\emptyset})\vert$}. }
    \label{fig:test_q}
\end{figure*}

Fig~\ref{fig:test_q} shows that the all AND-OR interactions encoded in three different {\small $\kappa$} value settings are almost the same. This indicates that the extraction of interactions is relatively robust to the {\small $\kappa$} value.

\section{Discussions on Distinctive Information-Processing Behaviors of Each Specifc DNN}
\label{app_sec:discuss_distinc}
We discovered that in most DNNs, low layers and middle layers usually learned to fit target interactions that were finally used by DNNs at the cost of encoding lots of redundant interactions. Such redundant interactions would be removed in high layers.

Distinctive information-processing behaviors of different DNNs. Specifically, for DNNs trained on the MNIST dataset, they usually learned the target interactions for inference quickly, because
the MNIST dataset was easy to learn. Particularly, for the ResNet-20 trained on both the MNIST dataset and the CIFAR-10 dataset, its low layers and middle layers mainly learned target interactions for inference, while high layers mainly forgot high-order interactions. These high-order interactions were unstable and exhibited poor generalization capacity, as verified in Section 3.3.

For MLP-7 and VGG-11 trained on the CIFAR-10 dataset, low layers were unable to learn interactions that could be directly used for classification, due to the challenge of classification on the CIFAR
dataset. Then, middle layers and high layers gradually learned the target interactions for inference without generating redundant interactions. High layers did not change the interactions significantly.

For the DistilBERT and BERTBASE trained on the SST-2 dataset, low layers usually could not encode target interactions. Then, middle layers gradually learned the target interactions for inference, but also brought in lots of redundant interactions. High layers usually forgot redundant interactions, which were mainly high-order and unstable.

\section{Discussion on the Noise Ratio of Interactions over Different Orders}
\label{app_sec:Signal_to_noise}
In section 3.3 of the main paper, we add a small Gaussian perturbation {\small $\boldsymbol{\epsilon} \sim \mathcal{N}(\boldsymbol{0}, \sigma^2\boldsymbol{I})$} to the input sample {\small$\boldsymbol{x}$}, in order to mimic inevitable noises/variations in data. And then we use the new metrics to measure the relative stability of AND-OR interactions of each order {\small $m$} as follow.

\begin{equation}\begin{small}\begin{aligned}
\textit{\text{stability}}^{(l),m}_{\text{and}} ={\mathbb{E}}_{\boldsymbol{x}}\mathop{\mathbb{E}}_{S\in\Omega_{\text{and}}^{(l),m}} \!\!\left[ \vert E^{(l),m}_{\text{and}}(S,\boldsymbol{x})\vert \bigg/ \sqrt{\textit{Var}^{(l),m}_{\text{and}}(S,\boldsymbol{x})}\  \right]\\
\textit{\text{stability}}^{(l),m}_{\text{or}} = {\mathbb{E}}_{\boldsymbol{x}}\mathop{\mathbb{E}}_{S\in\Omega_{\text{or}}^{(l),m}} \!\!\left[ \vert E^{(l),m}_{\text{or}}(S,\boldsymbol{x})\vert \bigg/ \sqrt{\textit{Var}^{(l),m}_{\text{or}}(S,\boldsymbol{x})}\  \right]
\end{aligned}\end{small}\end{equation}

where {\small$ E^{(l),m}_{\text{and}}(S,\boldsymbol{x})=\mathbb{E}_{\boldsymbol{\epsilon}}[I_\text{and}(S \vert \boldsymbol{x} + \boldsymbol{\epsilon}, v^{(l)})]$} and 
{\small$\textit{Var}^{(l),m}_{\text{and}}(S,\boldsymbol{x})= \textit{Var}_{\boldsymbol{\epsilon}}[I_\text{and}(S \vert \boldsymbol{x} + \boldsymbol{\epsilon}, v^{(l)})]$} denote the mean and variance of the AND interaction {\small$I_\text{and}(S \vert \boldsymbol{x} + \boldsymbol{\epsilon}, v^{(l)})$}~\textit{w.r.t.} Gaussian perturbations {\small$\boldsymbol{\epsilon}$}, which are encoded by the $l$-th layer of the DNN. In fact, higher-order interactions comprise more input variables, which means it would obtain more gaussian perturbations. However, the signal strength of higher order interaction also increases linearly with order {\small $m$}. Fig.5 in the main paper shows that low-order interactions are more stable to small noises. Here we verify that the noise ratio of each interaction is relatively consistent over interactions of different orders, so that stability enables a fair comparison of interactions of different orders.

To this end, we used the images in CIFAR-10 dataset and annotated semantic parts in each image, following Appendix~\ref{app_sec:part}. Then we added the small Gaussian perturbation {\small $\boldsymbol{\epsilon} \sim \mathcal{N}(\boldsymbol{0}, \sigma^2\boldsymbol{I})$} to the image {\small $x$}, and obtained the noise image  {\small $x_\epsilon$}. For each interaction pattern {\small $S$}, we calculate the ratio of noise intensity to interaction pattern signal intensity, \textit{e.g.,} {\small $Radio^S_{\epsilon} = {\left\lVert x^s - x^s_\epsilon  \right\rVert_2}/{\left\lVert x^s\right\rVert_2}$}.

\begin{figure*}[ht]
	\centering
	\includegraphics[width=\linewidth]{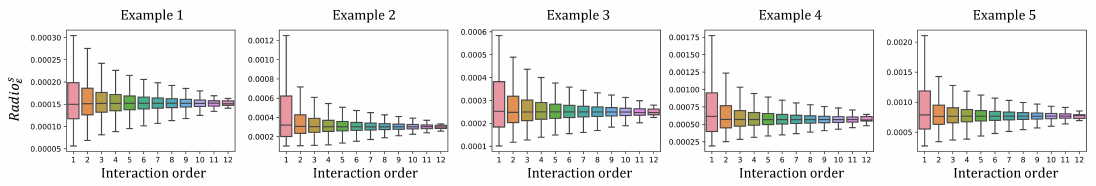}
	\caption{Box-and-whisker diagram of $Radio^S_{\epsilon}$ for interaction pattern $S$ of different interaction order.}
	\label{fig:ratio}
\end{figure*}

Fig.~\ref{fig:ratio} shows the distribution of {\small $Radio^S_{\epsilon}$} under different interaction order. It can be found that the median {\small $Radio^S_{\epsilon}$} values for different orders interactions is basically the same, and there is no phenomenon that the noise proportion of high-order interactions is significantly higher than that of low-order interactions. Therefore, the poor stability of the high order interaction is not due to its noise ratio.

\section{Experimental details}
\label{app_sec:exp_detail}

\subsection{Annotating Semantics Parts}
\label{app_sec:part}
We followed~\citep{pmlr-v202-li23at} to annotate semantic parts in MNIST dataset and CIFAR-10 dataset.
Given an input sample $\boldsymbol{x}\mathbb{R}^{n}$, the DNN may encode at most {\small$2^{n}$} interactions.
The computational cost for extracting salient interactions is high, when the number of input variables {\small$n$} is large. 
In order to overcome this issue, we simply annotate {\small$10\,\text{--}12$} semantic parts in each input sample, such that the annotated semantic parts are aligned over different samples in the same dataset. 
Then, each semantic part in an input sample is taken as a ``single" input variable to the DNN.

$\bullet$\;
For images in the MNIST dataset, we followed settings in~\citep{pmlr-v202-li23at}  to annotate semantic parts for {\small$100$} samples.
Specifically, given an image, we divided the whole image into small patches of size {\small$3\times3$}.
Considering the DNN mainly used the digit in the foreground to make inference, we selected {\small$n=10$} patches in the foreground as input variables to calculate interactions, in order to reduce the computational cost.

$\bullet$\;
For images in the CIFAR-10 dataset, we followed settings in~\citep{Ren_2023_CVPR} to annotate semantic parts for {\small$30$} samples.
Specifically, given an image, we divided the whole image into small patches of size {\small$4\times4$}, thereby obtaining {\small$8\times 8$} image patches in total.
Considering the DNN mainly used information contained in the foreground to make inference, we randomly selected {\small$n=12$} patches from {\small$6\times 6$} image patches located in the center of the image, in order to reduce the computational cost.

$\bullet$\;
For the SST-2 dataset, we followed settings in~\citep{Ren_2023_CVPR} to select sentences with a length of {\small$10$} words without unclear semantics, such as stop words.
For each selected sentence, we considered each word as an input variable, thereby obtaining {\small$n=10$} input variables in sum.
We used $50$ sentences to calculate interactions in Section~3.

\subsection{Training Linear Classifier in Section~\ref{sec:3.2.2}}
\label{app_sec:train}
Generally, the training parameters of the intermediate layers classifier are consistent with those of the original model.

$\bullet$\;
For MLP-7 trained on the MNIST dataset, we used SGD with learning rate 0.01, and set the batch size to 256 to train the intermediate layers. For  VGG-11 trained on both the MNIST dataset, we used SGD with learning rate 0.001, and set the batch size to 256 to train the intermediate layers. For  ResNet-20 trained on both the MNIST dataset, we used SGD with learning rate 0.001, and set the batch size to 256 to train the intermediate layers.

$\bullet$\;
For MLP-7 trained on the CIFAR-10 dataset, we used SGD with learning rate 0.001, and set the batch size to 256 to train the intermediate layers. For  VGG-11 trained on both the CIFAR-10 dataset, we used SGD with learning rate 0.001, and set the batch size to 256 to train the intermediate layers. For  ResNet-20 trained on both the CIFAR-10 dataset, we used SGD with learning rate 0.001, and set the batch size to 100 to train the intermediate layers.

$\bullet$\;
For DistilBERT finetuned on the SST-2 dataset, we used SGD with learning rate 2E-5, and set the batch size to 32 to train the intermediate layers. For BERTBASE finetuned on the SST-2 dataset, we used SGD with learning rate 1E-5, and set the batch size to 16 to train the intermediate layers.

\subsection{Intermediate Layers Selected to Calculate Interactions in Section~\ref{sec:representation}}
\label{app_sec:exp_detail_layer}

$\bullet$\;
For DNNs trained on both the MNIST dataset and the CIFAR-10 dataset, we used intermediate layers close to the output to compute interactions.
Specifically, 
the MLP-7 model contained $7$ linear layers, and we used features of the $4$-{th} linear layer.
For the VGG-11 model, we employed features of $conv4\_2$.
The ResNet-20 mdoel contained $9$ residual blocks, and we used features after the $6$-{th} residual block.
The ResNet-32 model contained  $15$ residual blocks, and we used used features after the $10$-{th} residual block.

$\bullet$\;
For DNNs trained on the SST2 dataset, we also used intermediate layers close to the output to compute interactions.
Specifically,
the DistilBERT model contained $6$ transformers, and we employed features after the $4$-{th} transformer.
The $\text{BERT}_\text{BASE}$ model contained $12$ transformers, and we employed features after the $8$-{th} transformer.
The XLNet model contained $12$ transformer-XLs, we employed features after the $8$-{th} transformer-XL.

\subsection{Experimental Details for Verifying the Sparsity of Interactions in Section~\ref{sec:3.2}.}
\label{app_sec:exp_sparsity}
For each sample in the MNIST dataset, as introduced in Appendix~\ref{app_sec:part}, we set $n=10$.
For each sample in the CIFAR-10 dataset, as introduced in Appendix~\ref{app_sec:part}, we set $n=12$.
We randomly selected $100$ images in the MNIST dataset and $30$ images in the CIFAR-10 dataset to verify the sparsity of interactions.
We set {\small $\tau=0.05 \cdot \max_{\boldsymbol{x}}\max_{S}( \max\{ \vert I_\text{and}(S\vert \boldsymbol{x},v^{(l)})\vert,$} {\small $ \vert I_\text{or}(S\vert \boldsymbol{x},v^{(l)})\vert \})$} for each target layer of the target DNN to determine its salient interactions.
Note that in experiments, we concluded first-order OR interactions to the first-order AND interactions for convenience.
In other words, the first-order AND interactions were the sum of first-order OR interactions and the first-order AND interactions, because one single input variable could be considered as either OR relationship or AND relationship with itself.


\end{document}